\documentclass[10pt,twocolumn,letterpaper]{article}

\usepackage{cvpr}              

\usepackage{xcolor}

\usepackage{multirow}
\usepackage{colortbl}
\usepackage{xcolor}
\usepackage{pifont} 

\definecolor{cvprblue}{rgb}{0.21,0.49,0.74}
\newcommand{\cmark}{\textcolor{green}{\ding{51}}} 
\newcommand{\xmark}{\textcolor{red}{\ding{55}}}   
\def\paperID{0000} 
\def\confName{xxxx}
\def\confYear{xxxx}
 \usepackage[pagebackref,breaklinks,colorlinks,allcolors=cvprblue]{hyperref}
\usepackage[capitalize]{cleveref}
\crefformat{section}{\S#2#1#3} 
\crefformat{subsection}{\S#2#1#3}
\crefformat{subsubsection}{\S#2#1#3}

\usepackage[T1]{fontenc}
\usepackage{amsfonts} 

\newcommand\extrafootertext[1]{%
    \bgroup
    \renewcommand\thefootnote{\fnsymbol{footnote}}%
    \renewcommand\thempfootnote{\fnsymbol{mpfootnote}}%
    \footnotetext[0]{#1}%
    \egroup
}

\title{UniMed-CLIP: Towards a Unified Image-Text Pretraining Paradigm \\ for Diverse Medical Imaging Modalities}

 \author{
  Muhammad Uzair Khattak$^{1, 2*\dag}$ \quad 
  Shahina Kunhimon$^{1*}$ \quad \\
  Muzammal Naseer$^{3}$  \quad
  Salman Khan$^{1,4}$ \quad
  Fahad Shahbaz Khan$^{1,5}$
  \vspace{0.2em} \\
  $^{1}$Mohamed bin Zayed University of AI \quad 
  $^{2}$EPFL \quad \\
    $^{3}$Khalifa University \quad 
 $^4$Australian National University 
\quad $^5$Link{\"o}ping University
}

\begin{document}
\maketitle
\begin{abstract}
Vision-Language Models (VLMs) trained via contrastive learning have achieved notable success in natural image tasks. 
However, their application in the medical domain remains limited due to the scarcity of openly accessible, large-scale medical image-text datasets.
Existing medical VLMs either train on closed-source proprietary or relatively small open-source datasets that do not generalize well. 
Similarly, most models remain specific to a single or limited number of medical imaging domains, again restricting their applicability to other modalities.
To address this gap, we introduce UniMed, a large-scale, open-source multi-modal medical dataset comprising over 5.3 million image-text pairs across six diverse imaging modalities: X-ray, CT, MRI, Ultrasound, Pathology, and Fundus. 
UniMed is developed using a data-collection framework that leverages Large Language Models (LLMs) to transform modality-specific classification datasets into image-text formats while incorporating existing image-text data from the medical domain, facilitating scalable VLM pretraining. 
Using UniMed, we trained UniMed-CLIP, a unified VLM for six modalities that significantly outperforms existing generalist VLMs and matches modality-specific medical VLMs, achieving notable gains in zero-shot evaluations. For instance, UniMed-CLIP improves over BiomedCLIP (trained on proprietary data) by an absolute gain of +12.61, averaged over 21 datasets, while using 3$\times$ less training data. To facilitate future research, we release UniMed dataset, training codes, and models at \href{https://github.com/mbzuai-oryx/UniMed-CLIP}{https://github.com/mbzuai-oryx/UniMed-CLIP.}

\end{abstract}

\extrafootertext{\textsuperscript{*}Joint first authors.  \textsuperscript{\dag} Work done while at MBZUAI.}

\section{Introduction}
\label{sec:intro}

Contrastive Vision-Language Models (VLMs) have significantly advanced multi-modal representation learning. Notable VLMs like CLIP \cite{radford2021learning} and ALIGN \cite{jia2021scaling} employ a self-supervised approach to jointly model the visual and textual data using a dual encoder architecture. 
By leveraging large-scale image-text pairs, VLMs learn a shared representation space, that enables their robust zero-shot and few-shot generalization performance across a multitude of tasks, including image recognition \cite{pratt2023does, menon2022visual, khattak2024learning}, segmentation \cite{zhou2022extract}, and retrieval \cite{tang2023lemons, radford2021learning, fang2023eva}. 
Having large-scale and diverse pretraining data has been a key factor in the effective adaptability and generalizability of these foundation models \cite{gadre2024datacomp, xu2023demystifying}.


\begin{figure}[tp]
  \centering 
    \includegraphics[width=0.90\linewidth]{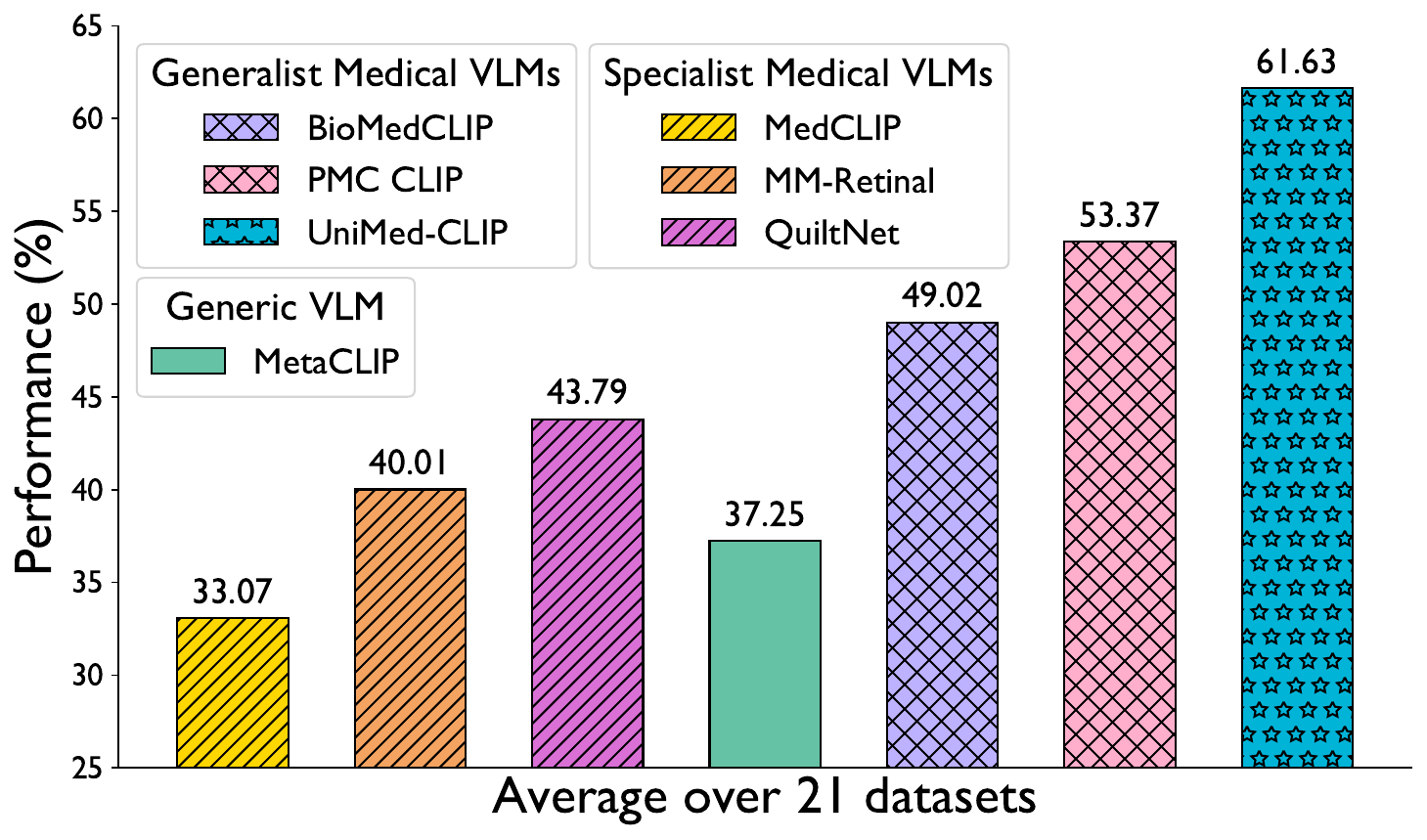}
    \caption{\textbf{Zero-shot medical image recognition results.} Averaged results over 21 datasets from 6 modalities: CT, MRI, US, X-ray, Histopathology, and Retinal fundus. 
    UniMed-CLIP trained on our open-source UniMed dataset developed using publicly available data sources 
    shows notable gains, compared to existing medical contrastive VLMs including MedCLIP \cite{wang2022medclip}, MM-Retinal \cite{wu2024mm}, QuiltNet \cite{ikezogwo2024quilt}, BiomedCLIP \cite{zhang2023biomedclip} and PMC-CLIP \cite{lin2023pmc}. }
  \label{fig:concept}
\end{figure}

The availability of millions of image-text pairs has become critical for effective contrastive learning in VLMs, where both the scale and quality of the dataset significantly impact model's performance \cite{fang2023eva, xu2023demystifying}. 
Studies have shown that carefully curated and larger datasets lead to richer multi-modal representations. This underscores the critical role of data-centric approaches in the development of contrastive VLMs. Consequently, recent works on VLMs have focused on refining pretraining datasets through quality filtering \cite{fang2023data, pouget2024no, gadre2024datacomp}, enriching captions \cite{fan2024improving, lai2023scarcity, hammoud2024synthclip, vasu2024clip}, and downstream task-aware pretraining \cite{naeem2023silc} while placing relatively less emphasis on architectural changes, as data quality has emerged as the main driver for performance in transformer-based foundational models \cite{peebles2023scalable,touvron2023llama,brown2020language}.

While general-purpose VLMs have made remarkable progress in the natural images domain, their potential applicability to the medical domain remains limited. The primary challenge in building Medical VLMs is the scarcity of openly accessible paired medical image-text data, which is crucial for effective contrastive learning in VLMs. Unlike natural image-text pairs that can be readily scraped from the internet \cite{xu2023demystifying, gadre2024datacomp}, curation and collection of medical datasets is significantly challenging due to the protected nature of medical data and privacy concerns \cite{price2019privacy, williamson2024balancing}. Additionally, most data-centric research contributions in the medical domain remain relatively inaccessible \cite{lu2024multimodal, zhang2023biomedclip, zhao2024biomedparse} to the wider research community for several potential reasons, including protected data agreements, competitive advantage, and potential applications in the healthcare market. This can potentially limit research advancements in the medical domain, especially for medical VLM pre-training. 
Medical data is inherently multi-modal and encompasses various types of medical imaging (e.g., radiology, retinal fundus, pathology) and corresponding textual data (e.g., clinical narratives, reports, annotations).
Therefore, medical data-centric research is both promising and impactful for developing the next-generation multi-modal foundational models for effective healthcare applications. 


In the medical literature, recent works have curated datasets for pretraining VLMs for various downstream medical tasks. 
MedCLIP \cite{wang2022medclip} uses image-text dataset for X-ray representation learning, while BiomedCLIP \cite{zhang2023biomedclip} curates the closed-source PMC-15M dataset to improve zero-shot and few-shot transfer. 
 Quilt-1M \cite{ikezogwo2024quilt} and MM-Retinal \cite{wu2024mm} train specialized foundation models using pathology and retinal datasets, respectively. 
  Despite these advancements, several key limitations remain unaddressed: (i) \textbf{Closed-source Datasets:} High-performing VLMs like BiomedCLIP \cite{zhang2023biomedclip} rely on proprietary datasets, which hinder the data-centric research due to lack of public access. (ii) \textbf{Small Scale:} VLMs such as MedCLIP \cite{wang2022medclip}, PMC-CLIP \cite{lin2023pmc}, MM-Retinal \cite{wu2024mm} and GLoRIA  \cite{huang2021gloria} are trained on small-scale datasets which limit their performance. (iii) \textbf{Modality Specific VLMs:} Most VLMs \cite{wu2023medklip, wu2024mm, silva2024foundation, ikezogwo2024quilt} focus on single modality, which limits their cross-modal generalization.

To address these gaps, we introduce UniMed, a large-scale and open-source medical multi-modal dataset created through a scalable data-collection framework. 
Addressing the scarcity of publicly available medical multimodal data, UniMed combines image-text and image-label datasets in a unified manner to scale the data for effective VLM pretraining. Specifically, we leverage Large Language Models (LLMs) \cite{gpt4o} to convert high-quality image-label data into image-text pairs, which, when combined with existing medical VLM data, yields 5.3 million image-text pairs across six diverse modalities: X-ray, CT, MRI, ultrasound, retinal fundus, and histopathology (see Tab. \ref{tab:ours_vs_others}).
Based on our UniMed dataset, we train contrastive VLM models that show significantly improved performances against existing generalist VLMs and are competitive with modality-specific VLMs. 
Our UniMed-CLIP achieves a gain of 12.61\% over BiomedCLIP \cite{zhang2023biomedclip} and 8.26\% over PMC-CLIP \cite{lin2023pmc} on zero-shot transfer averaged across 21 datasets (see Fig.~\ref{fig:concept}).  

\definecolor{verylightgray}{rgb}{0.97, 0.97, 0.97}

\begin{table}[t]
    \centering
    \setlength{\tabcolsep}{3pt}
    \resizebox{0.47\textwidth}{!}{%
    \begin{tabular}{lccccccc}
    \hline
    \textbf{Dataset Characteristics} & \textbf{MedCLIP} & \textbf{MMRetinal} & \textbf{Quilt} & \textbf{PMC-OA} & \textbf{PMC-15M} & \textbf{UniMed} \\     & \cite{wang2022medclip} & \cite{wu2024mm} & \cite{ikezogwo2024quilt}  & \cite{lin2023pmc} &  \cite{zhang2023biomedclip} &\small\emph{(ours)}
    \\    
    \hline
   Public \& Open-Source Datasets & \cmark & \cmark & \cmark & \cmark & \xmark & \cmark \\
      Training Code & \xmark & \cmark & \xmark & \cmark & \xmark & \cmark \\
    High-Quality image-only datasets & \xmark & \xmark & \xmark & \xmark & \xmark & \cmark \\
    \hline
     \rowcolor{verylightgray}
    \# Explicit Modalities & 1 & 1 & 1 & - & -& 6 \\
    \rowcolor{verylightgray}  \# Training samples & 0.57M & 0.18M & 1.1M & 1.6M & 15M & 5.28M  \\
    \bottomrule
    \end{tabular}}

    \caption{Comparison of UniMed with prior medical VLM pretraining datasets/Models. UniMed strives to be a completely open-source dataset and covers 6 diverse medical modalities incl. CT, MRI, Ultra Sound, Retinal Fundus, X-ray, and Histopathology.}
    \label{tab:ours_vs_others}
    \vspace{-1em}
\end{table}

\noindent To summarize, our main contributions are as follows:
\begin{itemize}
    \item We introduce UniMed: an open-source, large-scale multi-modal dataset developed using an LLM-in-the-loop framework, comprising over 5.3 million samples. 
    It covers six diverse medical modalities and provides a robust foundation for training generalizable medical VLMs.
    \item Building upon UniMed, we train a contrastive VLM, UniMed-CLIP, which is tailored to the medical domain and achieves impressive zero-shot results across various benchmarks spanning six medical imaging modalities.
    \item We perform extensive experiments showing significant improvements with UniMed-CLIP and report detailed ablation studies to validate our design choices. Our training code, dataset, and model checkpoints will be open-sourced to encourage further progress in  medical VLMs.
\end{itemize}

\section{Related Work}
\label{sec:related-work}

\noindent\textbf{Contrastive Vision-Language Models (VLMs).}
Contrastive VLMs  \cite{radford2021learning,jia2021scaling} have shown the effectiveness of large-scale paired image-text datasets for multi-modal learning, and achieves robust zero-shot transfer in classification and retrieval tasks. Building on this, VLMs like Florence \cite{yuan2021florence}, CoCa \cite{yu2022coca} and EVA-CLIP \cite{fang2023eva} have further optimized contrastive learning using larger datasets and refined training techniques. However, these VLMs are predominantly trained on natural images, and their downstream adaptation for medical imaging has proven challenging due to domain-specific shifts and complex medical vocabulary \cite{taghibakhshi2021optimization, wang2022medclip}.
In this work, we aim to advance medical VLM training and develop a generalist VLM trained entirely on public medical datasets converted to a multi-modal format suitable for VLM pretraining.

\begin{figure*}[t!]
  \centering
  \includegraphics[width=\linewidth]{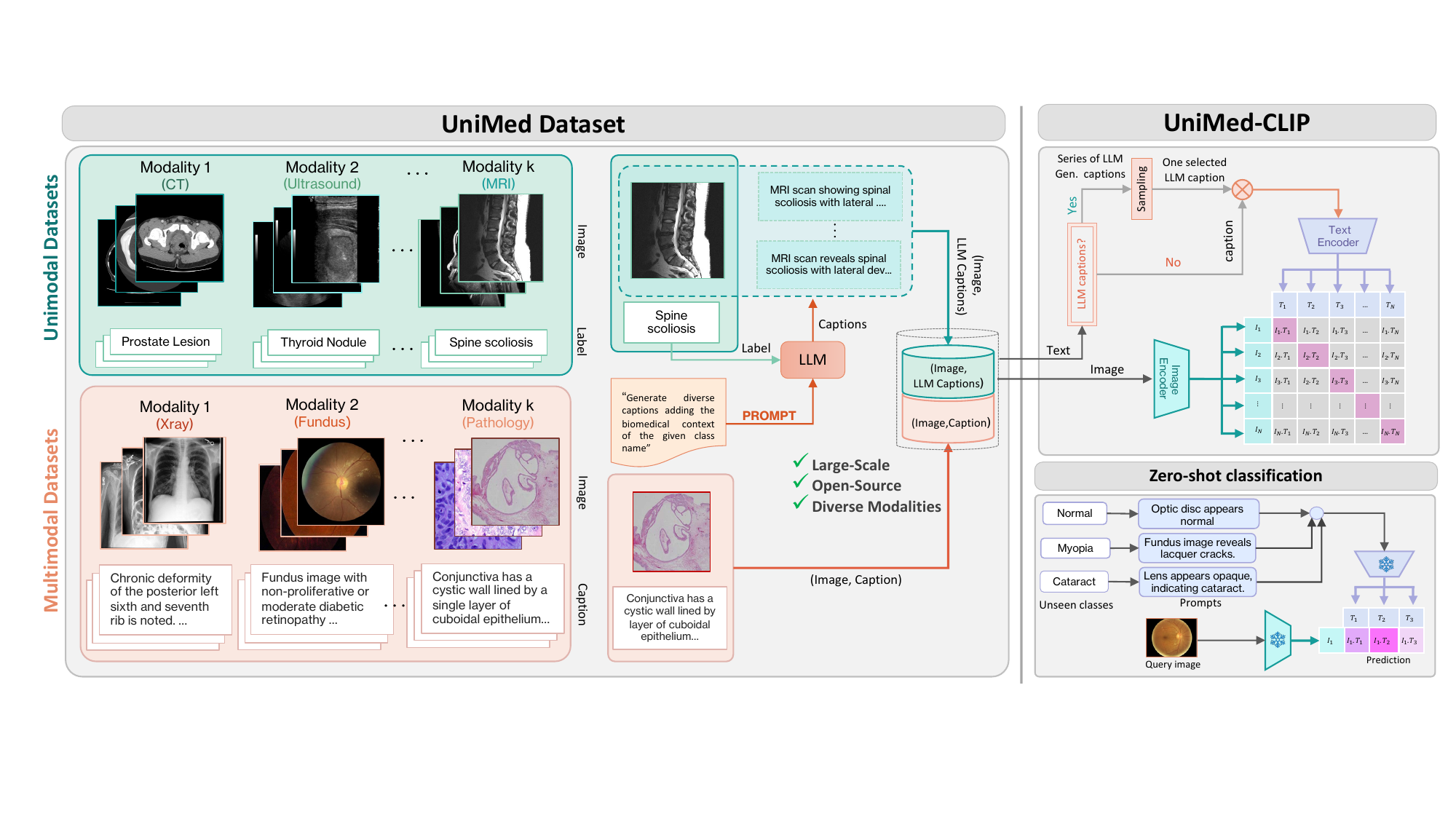}
 \vspace{-22 mm}
  \caption{\textbf{Overview of UniMed dataset and UniMed-CLIP VLM.} {\color{blue}\textbf{(Left)}}: We develop a medical pretraining dataset, UniMed by meticulously collecting publicly available label-only (uni-modal) image datasets and image-text (multi-modal) datasets. {\color{blue}\textbf{(Middle)}}: We utilize LLM-in-the-loop framework to convert label-only datasets into pseudo-image-text pairs where each image is paired with multiple captions. Both pseudo-image-text pairs and already available image-text pairs are used to create the UniMed dataset, which is a) open-source, b) large-scale and, c) covers diverse medical modalities. {\color{blue}\textbf{(Right)}}: Using UniMed dataset, we train UniMed-CLIP within a contrastive language-image pretraining paradigm. The resulting VLM performs well in zero-shot evaluations across various medical modalities.}
  \label{fig:dataset_curation}
\end{figure*}

\noindent\textbf{Medical VLMs.} 
Medical-specific VLMs have evolved to address the limitations of general-purpose models. Early works like ConVIRT \cite{zhang2022contrastive} and GLoRIA \cite{huang2021gloria} introduced contrastive learning for chest X-rays by aligning images with text reports. While effective, these models relied solely on paired data, limiting scalability. 
MedCLIP \cite{wang2022medclip} addressed this by incorporating both paired and unpaired datasets through a decoupled framework and semantic matching, though its focus remained on X-rays. 
Models like MedViLL \cite{moon2022multi}, PubMedCLIP \cite{eslami2023pubmedclip}, and MedKLIP \cite{wu2023medklip} further refined text-based supervision with domain-specific resources, yet most models continue to focus on single modalities, such as QuiltNet \cite{ikezogwo2024quilt} and PLIP \cite{huang2023visual} for pathology, or MM-Retinal \cite{wu2024mm} and FLAIR \cite{zhou2023foundation} for retinal imaging, which limits broader applicability. 
Attempts to develop generalist VLMs, such as BioMedCLIP \cite{zhang2023biomedclip} and PMC-CLIP \cite{lin2023pmc}, rely on large-scale biomedical image-text pairs scraped from scientific papers, which are often noisy and inconsistent. 
While some models \cite{lin2023pmc} provide open-source training code, others  \cite{zhang2023biomedclip} do not, which limits transparency and reproducibility. 
To this end, we train UniMed on public, multi-modal medical datasets, scaled using an LLM-in-the-loop framework, to enhance generalizability and reproducibility across a wide range of medical imaging tasks. \\
\noindent\textbf{Multi-modal training using uni-modal data.}
To address the scarcity of paired medical image-text data, recent medical VLMs have leveraged image-only datasets by generating captions through innovative techniques. 
MedCLIP \cite{wang2022medclip} uses a decoupled framework to incorporate both paired and unpaired data, while FLAIR \cite{zhou2023foundation}, tailored for retinal imaging, combines class labels with expert-driven descriptions to capture clinical relevance. BioViL \cite{boecking2022making} repurposes classification labels from chest X-rays as text proxies, structuring them into sentences via language models, and LLaVA-Med \cite{li2024llava} uses GPT-generated captions that blend classification labels with clinical context for improved accuracy. While effective, these works have focused on enhancing single-modality datasets, and often utilize fixed templates to convert data into a multi-modal format. 
In this work, we present an approach that utilizes both uni-modal and multi-modal datasets across diverse medical imaging modalities, supported by an LLM-in-the-loop framework to generate high-quality captions, enhancing the UniMed-CLIP's adaptability and robust performance in varied medical applications.

\section{Towards Open Medical Foundation Models}

Motivated by notable advancements in contrastive VLMs for natural images \cite{fang2023eva, gadre2024datacomp, naeem2023silc}, we aim to advance Medical VLMs pretraining, which is crucial for developing next-generation medical foundational models applicable for various tasks such as zero-shot medical disease recognition and efficient transfer to medical downstream datasets. A key requirement for high-performance VLMs is paired image-text data; however, such data is notably scarce and and rarely available in public medical datasets.  Currently, most large-scale datasets for medical domain pretraining are proprietary \cite{zhang2023biomedclip, zhao2024biomedparse}, which limits further research efforts.

In this work, we aim to narrow the aforementioned gaps in previous research by developing a fully transparent, large-scale, and diverse medical dataset to promote large-scale medical pretraining practices for developing performant medical contrastive VLMs.

To this end, we introduce the UniMed dataset that comprises 5.3 million image-text pairs obtained by carefully combining medical datasets from publicly available sources. 
Our UniMed dataset includes six dedicated medical imaging modalities (X-ray, CT, MRI, US, Pathology, and Fundus) to improve multimodality representation learning for \textit{unified} medical VLM pretraining. Finally, we use UniMed dataset to train a medical VLM, named UniMed-CLIP, that employs a multi-captioning based contrastive pretraining and achieves favorable performance across downstream tasks for six medical modalities.

We illustrate the UniMed dataset creation framework in Fig. \ref{fig:dataset_curation}. UniMed relies primarily on publicly available medical data and includes both label-only and image-text-based datasets. 
To compensate for the scarcity of medical image-text pairs, we collect different modality-specific image-label datasets and formulate a process that converts their label-only information into a corresponding textual description using a label-to-template-captioning approach based on Large-Language Models (LLM). 
This framework is scalable and enables us to create pseudo-image-text datasets from high-quality label-only medical datasets. 
Finally, we combine originally occurring image-text and pseudo-image-text datasets together that forms UniMed dataset.

Next, we present the UniMed construction pipeline. 
We first discuss the choice of data sources in \cref{lab:medical_datasets_collection} and \cref{lab:medical_datasets_collection_label_only}, followed by the multi-caption annotation generation process for label-only datasets in \cref{lab:medical_datasets_collection_label_only}. Finally, we perform a statistical analysis of UniMed in \cref{unimed_statistics} and conclude the method section with a discussion of UniMed-CLIP training in \cref{lab:unimed_clip_creation}.

\subsection{Curating medical image-text datasets}
\label{lab:medical_datasets_collection}
In order to construct high-quality and large-scale data for medical VLM pretraining, our first goal is to select a pool of datasets that (i) have high-quality labels, (ii) cover multiple medical image types to ensure diversity, and (iii) are open-source, to release our contributions to the community.

Specifically, we collect open-source datasets that are categorized into two groups: image-text datasets and image-only datasets that contain label-only annotations. Both categories have their own benefits and limitations. Image-text datasets provide rich multimodal information containing textual captions in free-form text format, which is ideal for VLM pretraining at scale. However, they are often noisy \cite{li2024if} and relatively scarce in the medical domain. On the other hand, label-only datasets are more abundant and contain more precise annotation information but are not inherently multimodal. We collect publicly available datasets from both categories covering both modality-specific and general-medical datasets for building our UniMed dataset.

\begin{figure}[t!]
  \centering
  \includegraphics[width=0.8\linewidth, trim=90 90 90 90, clip]{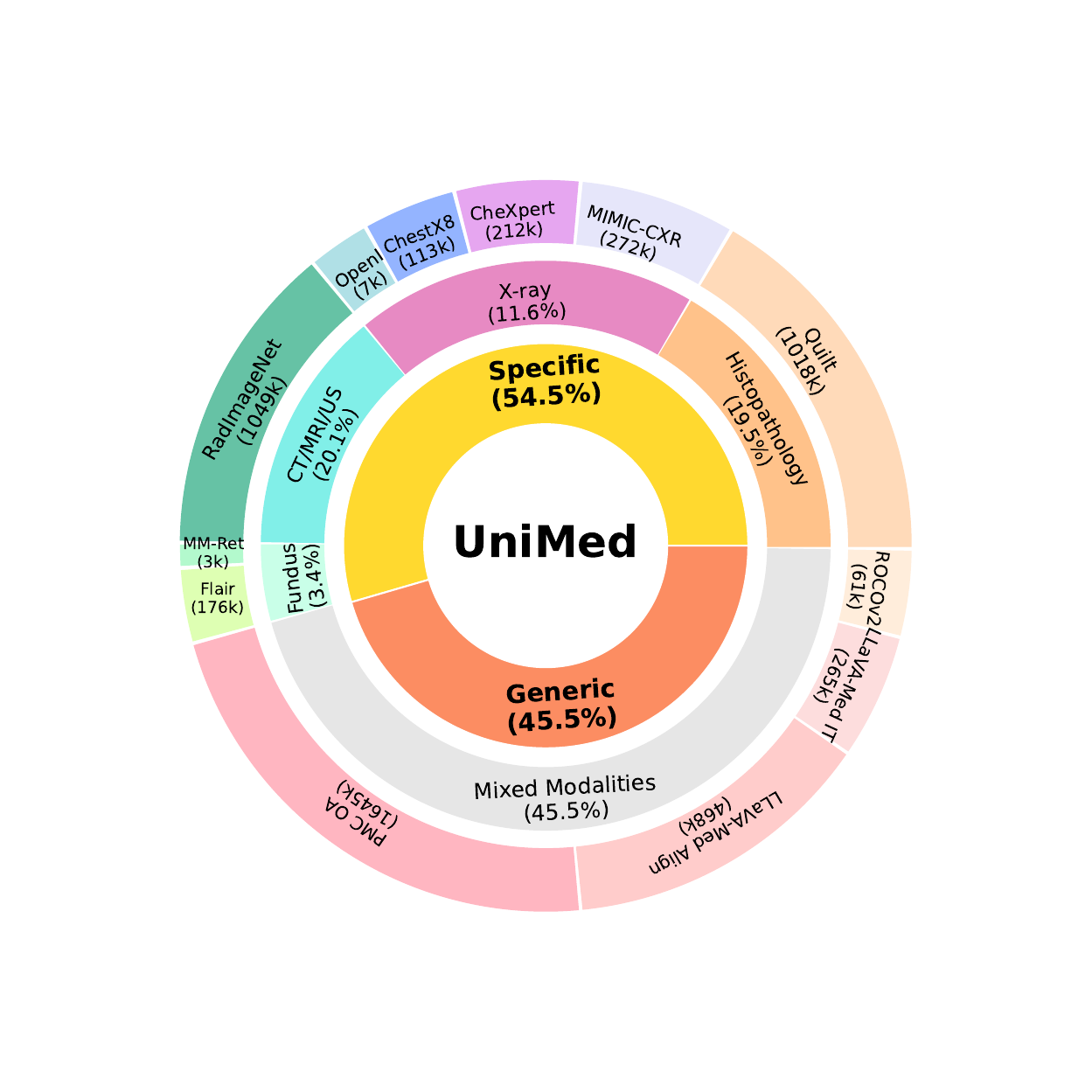}\vspace{-0.8em}
  \caption{\textbf{Public datasets used in UniMed:} It comprises both general medical domain datasets with diverse modalities and modality-specific datasets. The data spans publicly available unimodal (image-label) and multimodal (image-text) formats, enabling broad applicability across various medical imaging tasks.}
  \label{fig:overview_dataset}
\end{figure}

 \noindent\textbf{Modality-specific image-text datasets:} We collect publicly available modality-specific multimodal datasets. These include MM-Retinal  \cite{wu2024mm} for Retinal fundus modality, MIMIC-CXR \cite{johnson2019mimic} and OpenI \cite{demner2016preparing} for X-Rays. For the Histopathology modality, we utilize Quilt-1M dataset \cite{ikezogwo2024quilt}.
 
 \noindent\textbf{General medical image-text datasets:}
The majority of publicly available medical image-text datasets are scraped from scientific reports and research papers. In order to enhance the VLM training process, we leverage both the richness of scraped datasets like PMC-OA \cite{lin2023pmc} and ROCOV2 \cite{ruckert2024rocov2} for data diversity and the reliability of structured, high-quality image-label datasets which we discussed above.

Additionally, we note that image-text datasets used for fine-tuning Multi-modal LLMs, such as LLaVA-Med \cite{li2024llava}, can also be utilized for VLM pretraining. 
Therefore, we include their stage-1 image-text alignment data and inline mention captions data (stage 2) from LLaVA-Med \cite{li2024llava}. 
We associate the images with their corresponding inline mentions from the data and treat them as image-text pairs for the development of our UniMed dataset.

The details of the training samples for each collected dataset are shown in Fig. \ref{fig:overview_dataset}. We refer the readers to supplementary material (Sec. \ref{appendix:pretraining_datasets}) for additional details about the selected datasets and downloading instructions. 
 
\subsection{Curating image-only medical datasets}
\label{lab:medical_datasets_collection_label_only}
We further collect label-only medical datasets to address data-scarcity in medical VLM pretraining. Although the label-only datasets are not inherently multimodal, the category labels associated with images are of high quality, abundantly available, and cover diverse medical image types. 

\noindent\textbf{Modality specific datasets:} We collect RadImageNet \cite{mei2022radimagenet} that covers CT, MRI and US modalities, CheXpert \cite{irvin2019chexpert}, and Chest X-ray8 \cite{wang2017chestx} that encompasses X-ray modality, and FLAIR \cite{silva2024foundation} which consist of collection of datasets for the Retinal fundus medical modality. Next, we discuss the LLM-assisted multi-captioning technique to convert the image-label datasets into pseudo-image-text datasets.

\noindent\textbf{Label to Template Caption Generation:} \label{lab} 
In order to enable the effective  use of label-only datasets in VLM pretraining, we resort to generating template-based captions to convert the categorical labels into textual descriptions. 
For this, we explore LLMs \cite{gpt4o,jiang2023mistral,touvron2023llama} to generate captions for the label categories. 
Specifically, we provide the LLM with the \textbf{Label Info Triplet}, which includes the category label name or disease condition, anatomy type or organ (if available), and modality type of a given image, along with the prompt designed for label to template-caption generation. 
During training, each image is paired with a randomly selected caption from this set of multiple generated captions. 
This approach enhances caption \emph{diversity} while ensuring correct biomedical terminologies and class label information. 
While LLM-generated template captions do not provide localized label information within the image, the high-quality label information in these templates results in pseudo-image-text datasets that reinforce the originally available image-text data and provide complementary advantages during training. Further analysis is provided in the ablation study section (Sec. \ref{sec:Ablation}) and Appendix \ref{appendix:additional_tables}.

\begin{figure}[t!]
  \centering
  \includegraphics[width=\linewidth, ]{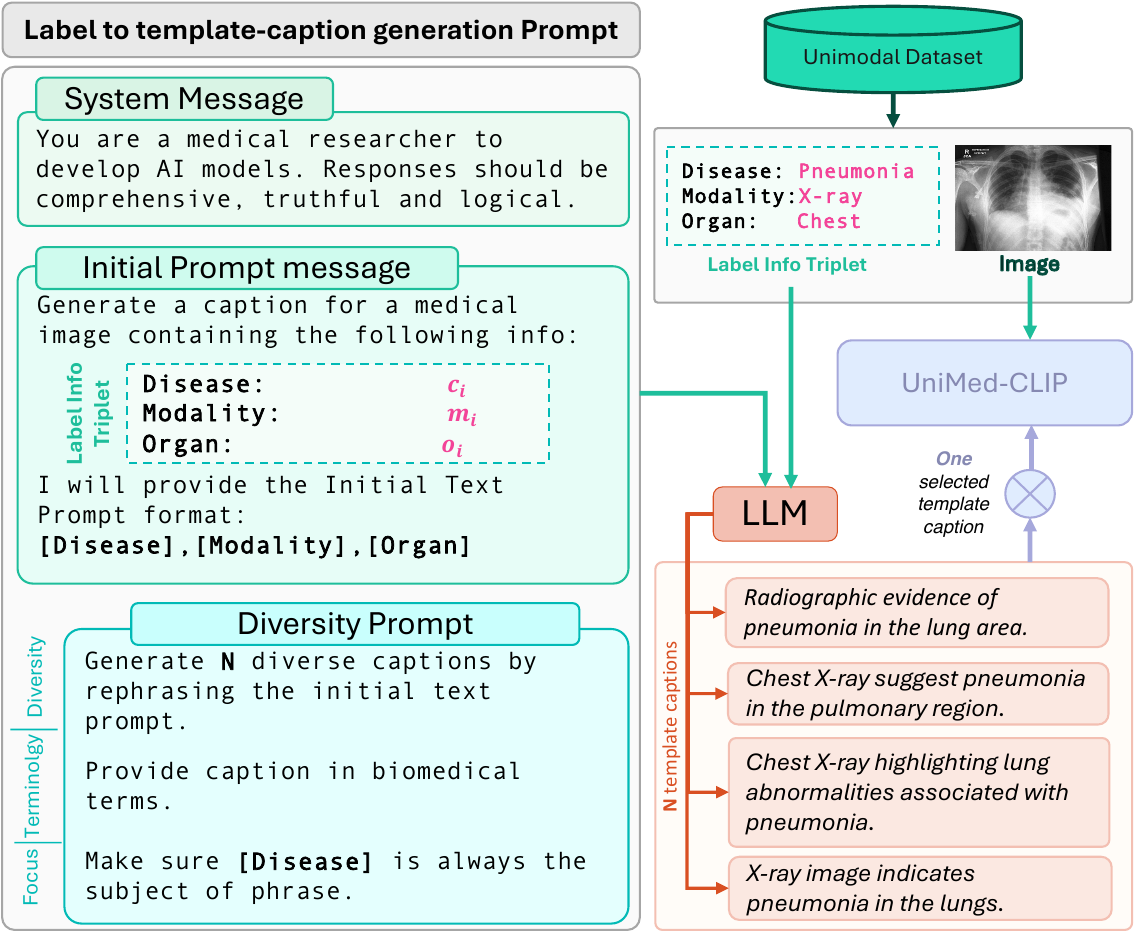}
\caption{\textbf{Label to Caption Generation Prompting:} We perform template caption generation using an LLM, which leverages available label information (\emph{Label Info Triplet}). This approach ensures diverse captions in agreement with ground-truth label information in biomedical terminologies. In each training iteration, each image is paired with a randomly sampled caption from its support set.}
\vspace{-1 em}
  \label{fig:prompt_generation}
\end{figure}

Specifically, given a set of label info triplets that contains category label, modality, and anatomy information $\{c_i, m_i, o_i\}_{i=1}^C $ extracted from label-only datasets, we prepare input prompts to LLM $\{L_{\texttt{inputs}}^i\}_{i=1}^C$ by inserting each triplet in a prompt formulated as:
\begin{equation}
L_{\texttt{inputs}}^i = \parbox{0.85\columnwidth}{
    \texttt{Generate a caption for a medical image containing the following information:} \\
    \texttt{Disease/Category Name: } \( c_i \), \\
    \texttt{ Modality Name: } \( m_i \), \\ 
    \texttt{ Organ Name: } \( o_i \)
}
\end{equation}

Qualitative template-caption examples using this prompt are shown in Fig.\ref{appendix: fig:example_captions} in supplementary material.

\noindent\textbf{Diversifying template-captions:}
\label{subsec:multi-label-captioning}
Although the aforementioned strategy generates plausible captions for label-only datasets, we observe that the generated captions have limited diversity and are similar to standard fixed templates, which can also be obtained using manual prompt engineering (e.g., A medical [modality] photo of a [Disease], in the [Anatomy]). Our early experiments with these captions showed sub-optimal performance in zero-shot evaluations. 

We overcome the lack of diversity in template captions using a two-fold approach. Firstly, we perform prompt engineering to steer the LLM to generate text captions in a medical professional style and nomenclature, similar to the texts that are found in naturally occurring medical image-text pairs and are written by professional medical doctors/researchers, such as the text captions of LLaVA-Med image-text dataset \cite{li2024llava}. Secondly, we instruct the LLM to produce multiple captions for the same triplet sample in diverse tones and styles. This results in having \textit{multiple} captions per category label. This process is illustrated in Fig. \ref{fig:prompt_generation}. During training, we randomly sample single caption from its corresponding set, which allows the occurrence of different captions with the same image during training. This strategy enforces further diversity in the label-only datasets and leads to better representation learning in medical VLMs.

Complete prompt message and additional qualitative results are shown in Fig. \ref {appendix: fig:sample_prompt} in Appendix \ref{appendix: fig:example_captions}. We observe that resulting template captions are diverse and close to free-form medical texts, which also improves its synergy with originally occurring image-text datasets during the joint training. We show ablations on these design choices in the analysis part of the paper in Sec. \ref{sec:Ablation}. Using this label-to-captioning technique, we convert all label-only datasets in our dataset pool. We note that this approach is scale-friendly and can be extended to additional image datasets. 

\begin{figure}[t!]
  \centering
  \includegraphics[width=\linewidth]{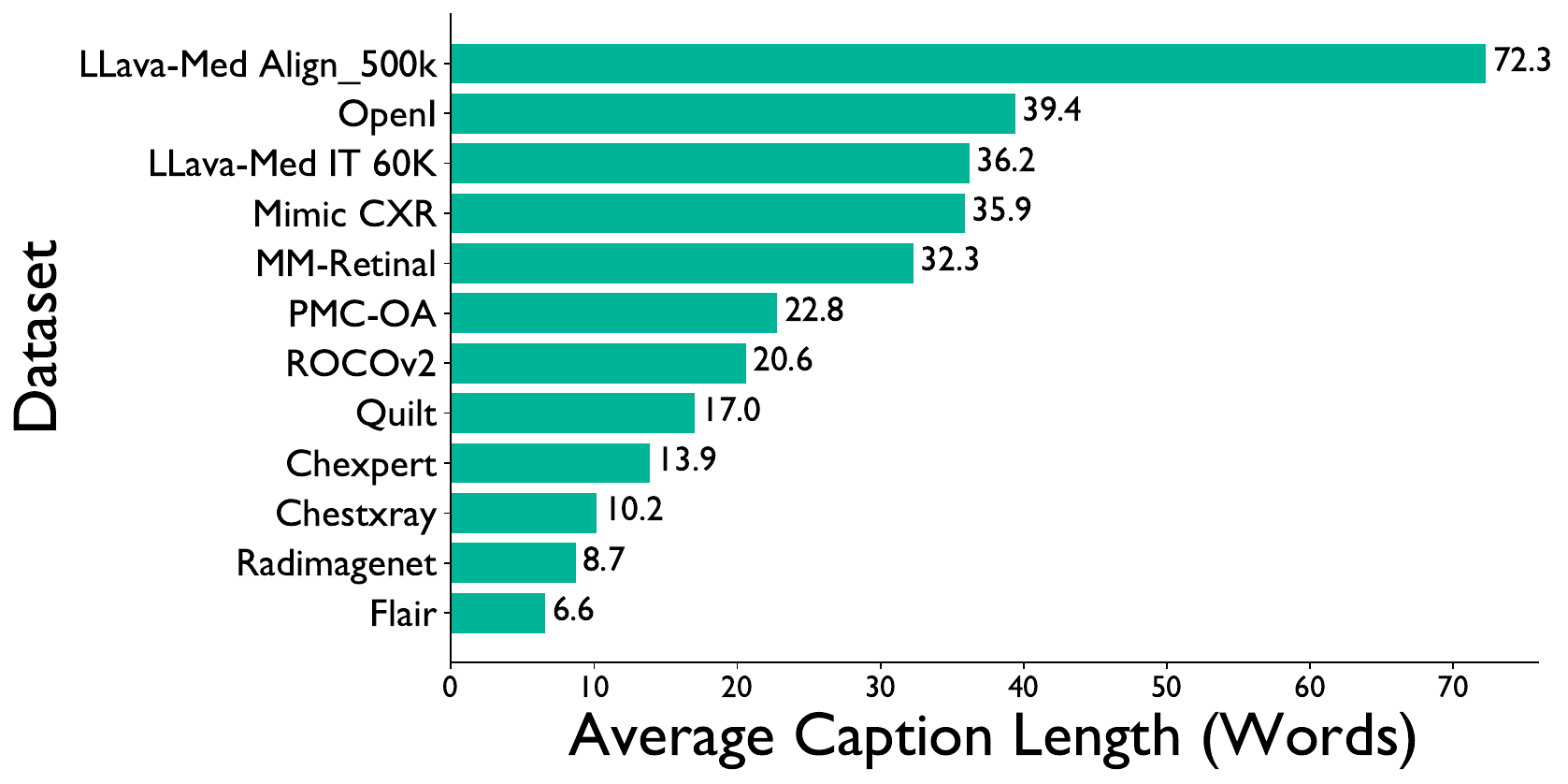}
   \caption{\textbf{ Caption length distribution:} Average length of captions (in words) of different datasets  used in UniMed.}

  \label{fig:textlength}
  \vspace{-1 em}
\end{figure}
\subsection{Distribution of UniMed dataset}
\label{unimed_statistics}

Overall, UniMed contains 5.3 million image-text pairs that covers six medical modalities. UniMed also covers additional modalities sourcing from the naturally occurring image-text datasets including ROCO, LLava-Med and PMC-CLIP. We show the statistical overview and contributions of each modality in UniMed dataset in Fig. \ref{fig:overview_dataset}. 

Histopathology stands as the most represented modality, contributing approximately 20\% in UniMed. X-ray follows with 12\% data contribution, supported using both image-text as well as pseudo-image-text data obtained from image-only datasets. CT, MRI, and Ultrasound together contribute approximately 20\% of the dataset. Fundus reaches around 3-4\% of the total contribution. The remaining 45.5\% constitutes the image-text pairs that come from the generic datasets containing various other modalities.

In Fig \ref{fig:textlength}, we analyze the average caption lengths across various data-sources of UniMed, including those generated using our label-to-template caption approach. UniMed contains data-samples with both short captions (E.g., Flair, RadImageNet etc) as well sa long-captions (e.g., OpenI and LLaVA-Med) which allows the model for learning rich semantics encompassing different caption lengths.

\subsection{Training UniMed-CLIP}
\label{lab:unimed_clip_creation}

Our main motivation for building the UniMed dataset is to train strong medical VLMs for advancing open-source medical imaging representation learning, aimed at solving a variety of downstream tasks. Building upon the UniMed as the pretraining dataset, we develop UniMed-CLIP medical VLM suitable for both zero-shot and adaptation tasks. 

For UniMed-CLIP architecture and training, we closely follow the Contrastive Language Image Pretraining (CLIP) paradigm \cite{radford2021learning}, with the primary adaptation being a multi-captioning strategy for label-only medical datasets (Sec. \ref{subsec:multi-label-captioning}). UniMed-CLIP employs separate image and text encoders. The pretraining objective uses a contrastive loss to maximize the cosine similarity of embeddings from aligned image-text pairs and minimize it for unaligned pairs in the joint vision-language embedding space.

For an aligned image-text dataset $D=\{I_i, T_i\}_{i=1}^{N}$, let $v_i = f_{\text{vision}}(I_i)$ represent the image embedding generated by passing image $I_i$ through the vision encoder $f_{\text{vision}}$. Similarly, let $t_i = f_{\text{text}}(T_i)$ denote the text embedding generated by passing the text $T_i$ through the text encoder $f_{\text{text}}$. 
The cross-entropy loss for aligning image-to-text and text-to-image similarities is defined as:

\[
L_{\text{i2t}} = -\frac{1}{2N} \sum_{i=1}^{N} \log \frac{e^{\text{sim}(v_i, t_i) / \tau}}{\sum_{j=1}^{N} e^{\text{sim}(v_i, t_j) / \tau}},
\]
\[
L_{\text{t2i}} = -\frac{1}{2N} \sum_{i=1}^{N} \log \frac{e^{\text{sim}(t_i, v_i) / \tau}}{\sum_{j=1}^{N} e^{\text{sim}(t_i, v_j) / \tau}},
\]
where $\text{sim}(\cdot, \cdot)$ denotes the cosine similarity, $\tau$ is the temperature parameter, and $N$ is the number of image-text pairs in the batch. The overall contrastive loss combines the image-to-text and text-to-image components:
\[
L_{\text{contrastive}} =  L_{\text{i2t}} + L_{\text{t2i}}.
\]
As discussed earlier, during training, we employ a multi-captioning strategy for samples in label-only datasets. Specifically, for an image-text pair in label-only category, we randomly select a single caption $T_{i}$ from a set of multiple captions $T\text{max}_i = \{T_{i1}, T_{i2}, \dots, T_{iM}\}$ associated with the image $I_i$. For samples from paired image-text datasets, we use the original single caption $T_i$ as the text input.

\begin{figure}
  \centering
  \includegraphics[width=\linewidth]{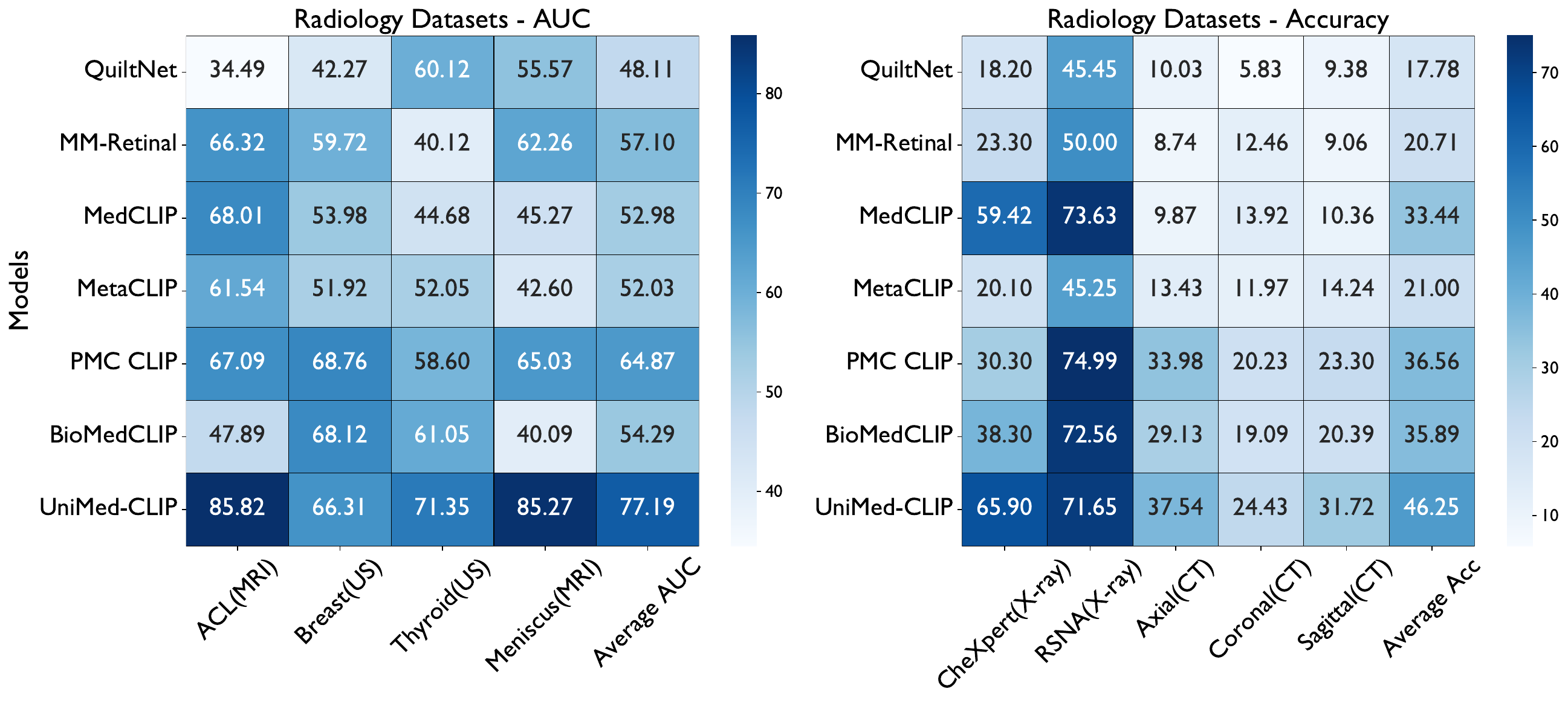}
  \vspace{-2em}
    \caption{\textbf{Zero-shot results on Radiology datasets:} Overall, UniMed-CLIP achieves improved average performance compared to both specialist and generalist medical VLMs.}
   \label{fig:HM_rad}
   \vspace{0em}
\end{figure}

\section{Experiments}
We perform zero-shot and downstream task transfer experiments to evaluate the performance of UniMed-CLIP, trained on the UniMed dataset, and show performance comparisons with existing medical VLMs. For zero-shot tasks, we evaluate models on 21 medical image recognition datasets that cover six different medical modalities. For the downstream adaptation task, we conduct linear probing experiments on 10 datasets to assess the suitability of frozen backbone representations for downstream tasks. Finally, we perform ablative analysis. We refer the readers to Appendix. \ref{appendix:Downstream_datasets} for details about datasets used in our evaluations. 
 
\textbf{Implementation details.}
All data sources used for developing UniMed are collected from their publically available repositories. In label to template caption generation component, we use OpenAI GPT4o \cite{gpt4o} model for generating pseudo captions. For UniMed-CLIP pretraining, we initialize its vision-encoder from MetaCLIP ViT-B/16 model \cite{xu2023demystifying}, and its text encoder is initialized from the BioMed-BERT text encoder \cite{chakraborty2020biomedbert} respectively. We fine-tune UniMed-CLIP for 10 epochs using 16 A100 40GB GPUs in a multi-node training setup. Learning rate is set to 5e-5 with a warmup phase of 2k iterations. Complete training of UniMed-CLIP takes 10 hours in total. Refer to supplementary material for additional implementation details.
\subsection{Zero-shot Medical Imaging Classification}
We present zero-shot evaluation experiments of UniMed-CLIP on 21 medical datasets grouped into 6 modality types and compare results with prior VLMs.  The selected datasets encompass different diagnostic tasks, such as disease detection, organ classification, grading, and tumor identification.
\noindent \textbf{Radiology.} For CT, MRI, X-ray, and ultrasound modalities, we present evaluation results in Fig. \ref{fig:HM_rad}. MedCLIP, which is an X-ray specialist VLM, shows improved performance on X-ray datasets including CheXpert and RSNA. In contrast, generalist VLMs trained on web-scraped image-text pairs, including PMC-CLIP and BioMedCLIP, provide improved performance across multiple radiology modalities. Compared to these models, UniMed-CLIP, which is trained on UniMed dataset, shows the overall best results and achieves the highest results in 7 out of 9 datasets. \\

\noindent \textbf{Retinal Fundus.} We show zero-shot evaluation comparisons for the retinal fundus modality are shown in Tab. \ref{tab:color-fundus-zeroshotresults}. MM-Retinal \cite{wu2024mm} is a specialist VLM explicitly trained on Retinal data, shows the best results on 3/4 datasets. In comparison with generalist VLMs, our UniMed-CLIP surpasses on 3 out of 4 datasets with the averaged gain of 2.69\% against the best performing PMC CLIP VLM \cite{lin2023pmc}. \\

\begin{figure*}
  \centering
    \includegraphics[width=\linewidth]{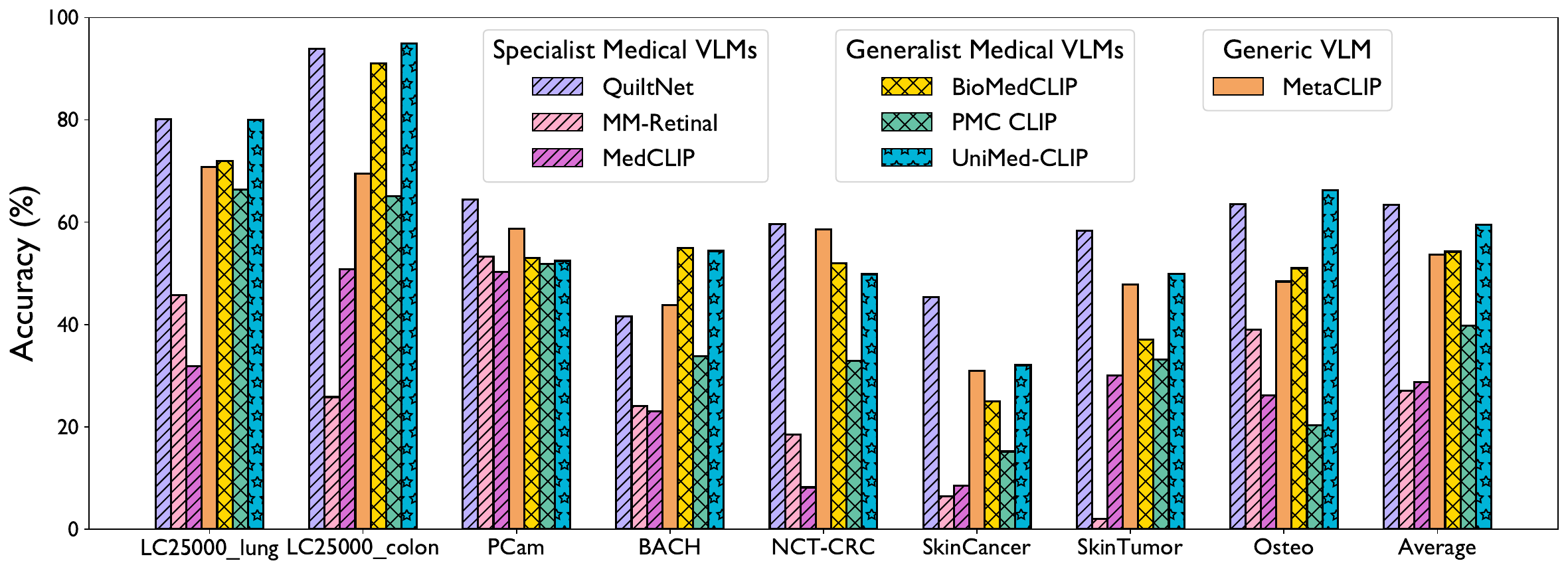}
  \caption{\textbf{Zero-shot performance on histopathology datasets:} UniMed-CLIP demonstrates performance comparable to the histopathology-specific QuiltNet \cite{ikezogwo2024quilt} and outperforms other generalistic medical VLMs in average performance (rightmost bars).}
  \label{fig:Histopath_Datasets_plot_plain}
   \label{fig:Histo_plot}
\end{figure*}
\begin{table}[t]
    \centering
    \scriptsize  
    \setlength{\tabcolsep}{3pt}  
    \begin{tabular}{l c c c c c}
        \toprule
        \textbf{Model} & \textbf{REFUGE} & \textbf{DR\_Reg} & \textbf{FIVES} & \textbf{ODIR 200x3} & \textbf{Average} \\
        & (AUC) & (ACC) & (ACC) & (ACC) & \textbf{} \\
        \midrule
        \multicolumn{6}{c}{\underline{\textbf{Specialist Models}}} \\
        MM-Retinal & \textbf{94.10} & \underline{45.00} & \textbf{73.10} & \textbf{81.20} & \textbf{66.43} \\
        MedCLIP & 38.99 & 14.50 & 19.13 & 33.16 & 22.93 \\
        Quilt 1M & 59.25 & 9.00 & 25.38 & 37.66 & 24.68 \\
        \midrule
        \multicolumn{6}{c}{\underline{\textbf{Generalist Models}}} \\
        MetaCLIP & 62.03 & 10.50 & 25.50 & 52.50 & 29.50 \\
        PMC CLIP & 68.99 & 43.75 & 57.13 & \underline{80.16} & 60.35 \\
        BioMedCLIP & 58.66 & 17.25 & 46.38 & 74.50 & 46.71 \\
        \midrule
        \rowcolor{cyan!15} UniMed-CLIP & \underline{85.45} & \textbf{51.75} & \underline{67.38} & 70.00 & \underline{63.04} \\
        \bottomrule
    \end{tabular}
    \caption{\textbf{Zero-shot evaluation results on Retina Fundus datasets.} UniMed-CLIP substantially improves on 3/4 datasets compared to prior generalist VLMs and is competitive to the MM-Retinal \cite{wu2024mm} Retinal modality specialist VLM. Best results are in bold, and the second-best results are underlined.}
    \label{tab:color-fundus-zeroshotresults}
    \vspace{-1em}
\end{table}
\noindent\textbf{Histopathology.} We perform zero-shot evaluations on 8 histopathology datasets and compare results with generalist and specialist VLMs in Fig.~\ref{fig:Histo_plot}. QuiltNet \cite{ikezogwo2024quilt} on average shows the highest performance across these datasets as it is pretrained solely on histopathology modality. MedCLIP, on the other hand, provides the lowest average performance of 23\%, suggesting that modality-specific VLMs have limited potential to perform reasonably on other out-of-domain modalities. Among the generalist VLMs, UniMed-CLIP shows the highest average performance of 58\% and surpasses BioMedCLIP and PMC-CLIP on 5 out of 8 datasets. The aforementioned zero-shot results across different modalities suggests that unification of image-text datasets and label-only datasets as employed by UniMed-CLIP serves a strong baselines for specialist models and shows robust generalization performance on unseen datasets. Refer to Appendix \ref{appendix:additional_tables} for the numerical results. 

\vspace{-0.5em}
\subsection{Downstream task transfer with Linear Probing}
\vspace{-0.5em}
To assess the transferability of the learned representations, we conduct linear probing experiments, where we freeze the pre-trained encoders and finetune a linear head on the downstream task data. This approach measures how well the pre-trained foundational model generalizes to new tasks with minimal task-specific fine-tuning.

\begin{figure}[ht]
  \centering
     \includegraphics[width=\linewidth]{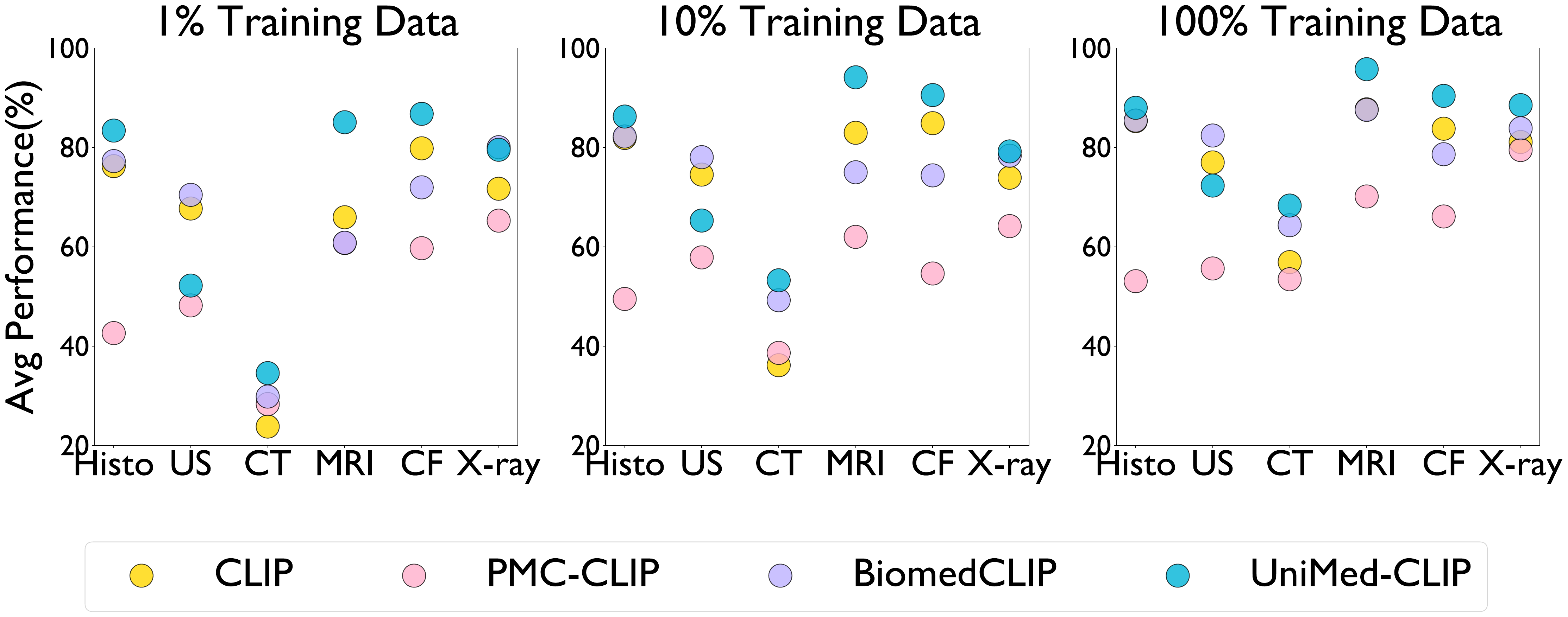}
  \caption{\textbf{Linear probing results:} Modality-wise averaged performance for linear probing. UniMed-CLIP shows strong performance, particularly when training data is limited.}

  \label{fig:LP_Bubble}
  \vspace{-1em}
\end{figure}

We perform linear probing experiments with the same datasets as used in zero-shot evaluations, using 1\%, 10\%, and 100\% of data points sampled from their respective training sets. We present the modality-wise averaged results in Fig. \ref{fig:LP_Bubble}. Moving from 1\% data towards 100\% data, UniMed-CLIP shows consistent improvements and surpasses previous generalist models on 5 out of 6 modalities. Overall, UniMed-CLIP achieves the highest gains for the MRI modality. With only 1\% of data points, UniMed-CLIP matches the performance of BioMedCLIP and PMC-CLIP trained on 100\% data on Histopathology, MRI, and Retinal Fundus modalities. This suggests that UniMed-CLIP has learned generalized representations suitable for multiple downstream medical modalities.

\section{Ablative Analysis}
\label{sec:Ablation}
\textbf{Effect of Textual Descriptions Diversity in image datasets.}
During the training of UniMed-CLIP, we use multiple captions for image-label datasets and randomly sample a single caption to enhance template-caption diversity. Here, we ablate on the number of template captions and study how the use of multiple captions affects the final performance. As shown in Tab. \ref{tab:LINEPLOT_AB3} (right), utilizing more templates leads to notable improvements in the zero-shot results, suggesting that multiple captions per image improve diversity in label-only datasets for VLM pretraining.

\begin{table}[t!]
    \centering
    \scriptsize  
    \setlength{\tabcolsep}{2pt}  
    \begin{tabular}{c c c c c c c c}
        \toprule
        \textbf{Templates} & \textbf{PCAM} & \textbf{Meniscus} & \textbf{CheXpert} & \textbf{Thyroid} & \textbf{Sagittal} & \textbf{FIVES} & \textbf{\cellcolor{cyan!15}Avg} \\
        \# & (Histo) & (MRI) & (X-ray) & (US) & (CT) & (Fundus) & \cellcolor{cyan!15}(\%) \\
        \midrule
        \textbf{1} & 50.56 & {86.57} & 52.90 & 60.93 & 28.15 & 36.93 & \cellcolor{cyan!15}52.67 \\
        \textbf{5} & {52.21} & \textbf{89.32} & {62.00} & {59.41} & {21.84} & {52.13} & \cellcolor{cyan!15}{56.15 }\\  
        \textbf{10} & \textbf{52.41} & 85.27 & \textbf{65.90} & \textbf{71.35} & \textbf{31.72} & \textbf{67.38} & \cellcolor{cyan!15}\textbf{62.33 }\\    
        \bottomrule
    \end{tabular}
    \caption{\textbf{Ablation on number of template-captions:} Comparison of performance when using a fixed single caption versus selecting from 10 caption templates for the same image for textual diversity.}
    \label{tab:LINEPLOT_AB3}
    \vspace{-1em}
\end{table}

\begin{figure}[t]
    \centering
    \begin{minipage}{0.49\linewidth}
        \centering
        \includegraphics[width=\linewidth]{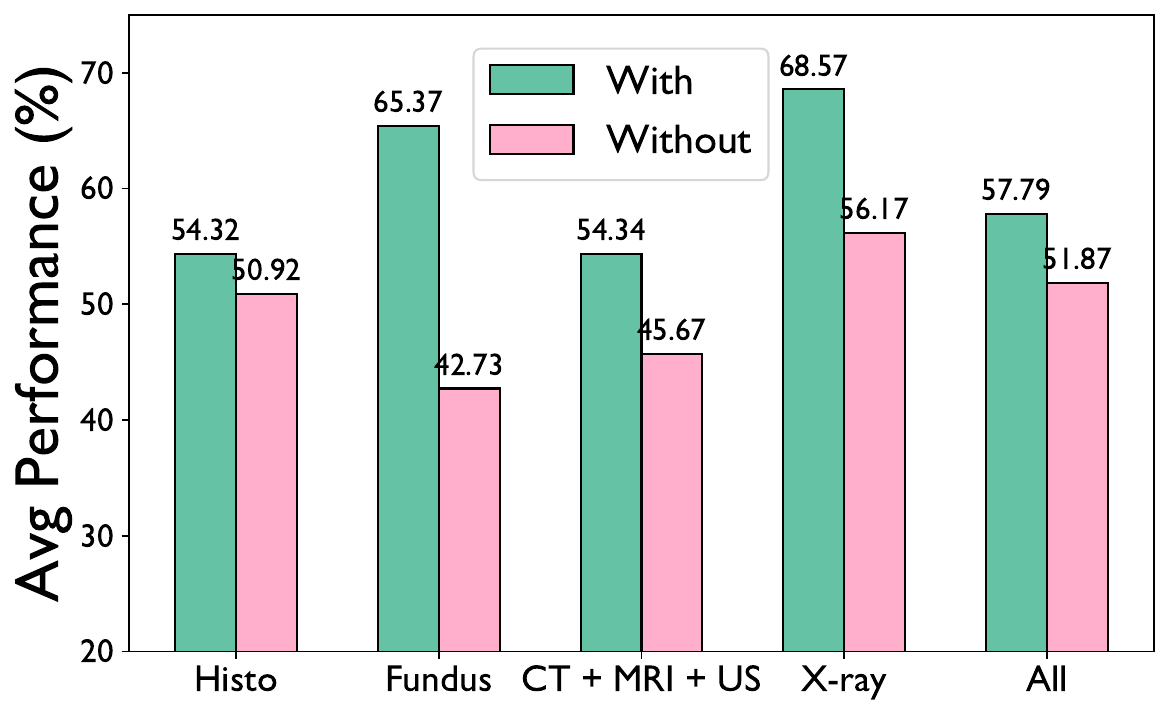}
        \label{fig:LINEPLOT_AB1}
    \end{minipage}
    \hfill
    \begin{minipage}{0.49\linewidth}
        \centering
        \includegraphics[width=\linewidth]{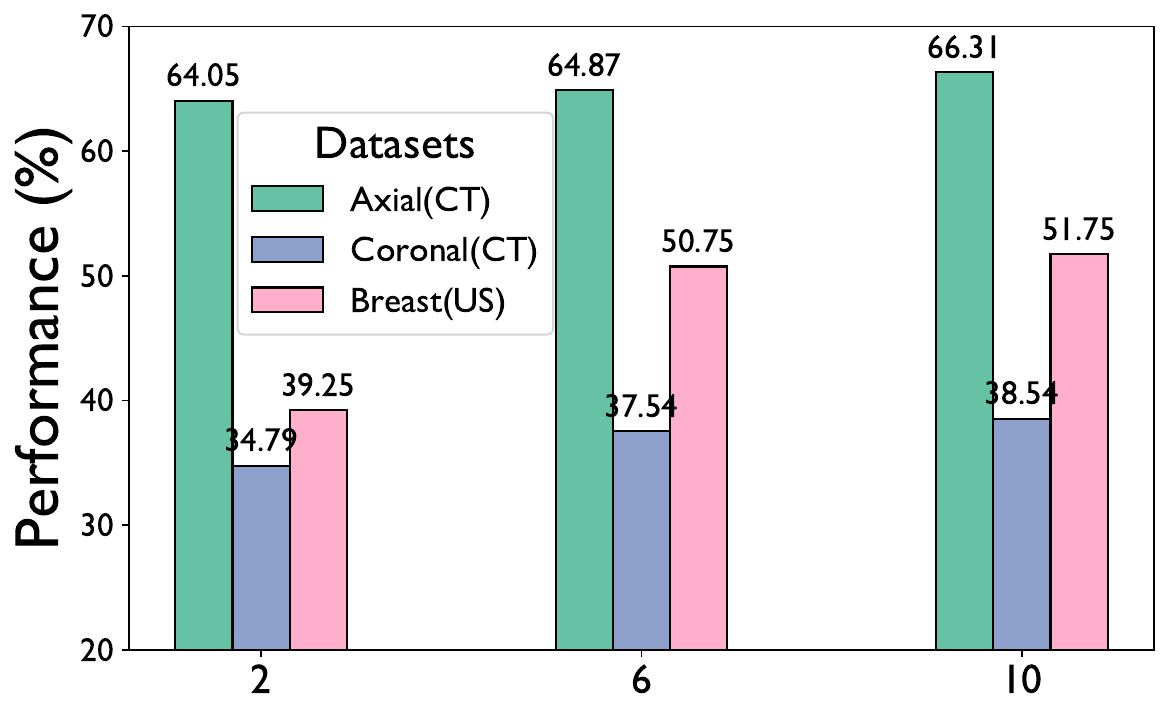}
        \label{fig:LINEPLOT_AB2}
    \end{minipage}
    \vspace{-2em}
    \caption{\textbf{(left) Contribution of modality-specific datasets for training UniMed-CLIP:} Dropping modality-specific datasets leads to a decline in average modality performance.\textbf{(Right) Effect of prompt ensembling at inference:} Increasing prompt templates leads to improved performance.}
    \label{fig:combined_ablation}
\end{figure}

\noindent \textbf{Test-time Prompt Ensembling.}
We next study the effect of textual prompt ensembling on the zero-shot performance of UniMed-CLIP in Fig. \ref{fig:combined_ablation}. We generate prompt templates from GPT-4o and perform prompt ensembling by varying the number of templates. Increasing prompt templates in UniMed-CLIP, leads to better zero-shot performance across different datasets. We refer the readers to supplementary material for qualitative examples of prompt templates.

\noindent \textbf{Importance of modality-specific data in UniMed-CLIP.}

We conduct an ablative analysis to gauge the contribution of modality-specific datasets in UniMed to UniMed-CLIP's performance. Specifically, we progressively remove modality-specific datasets, one at a time, train the VLM with the remaining data, and compare the performance results to the original base model, which is trained on the entire UniMed dataset. We present the results in Fig. \ref{fig:combined_ablation} (left).

The results reveal a clear trend: removing modality-specific datasets leads to a decline in performance for that specific modality, and this is the case for all modalities. This decline is particularly significant for certain modalities, such as retina and X-ray, indicating a strong dependence on modality-specific data for optimal performance. The overall performance (All) is also notably reduced when all modality-specific datasets are removed. This analysis emphasizes the importance of diverse and specialized training data for training performant generalist medical VLMs. 

\section{Conclusion}
Existing medical VLMs rely on closed-source data, limited dataset scale and/or remain specific to few modalities. 
We present UniMed, a large-scale, open-source, multimodal medical dataset specifically designed to address the limitations of existing medical VLMs. 
By leveraging a data-collection framework that converts high-quality image-label data into image-text pairs, UniMed provides 5.3 million pairs across six diverse medical imaging modalities (X-ray, CR, MRI, ultrasound, retinal fundus, and histopathology). 
We train a unified contrastive model on UniMed, called UniMed-CLIP that achieves significant performance gains, outperforming generalist VLMs and matching modality-specific models in zero-shot evaluations. 
We will publicly release UniMed, along with code and models, to drive further advancements in the medical VLM domain.
{
    \small
    \bibliographystyle{ieeenat_fullname}
    \bibliography{main}
}

 \clearpage
\setcounter{page}{1}
\appendix
\maketitlesupplementary

The following sections provide supplementary material for our main paper. This includes additional information on pretraining and downstream transfer Datasets, implementation details, and additional experimental results. The contents are organized as follows:
\begin{itemize}

        \item Dataset sources for UniMed (Section~\ref{appendix:pretraining_datasets})
        \item Downstream Datasets (Section~\ref{appendix:Downstream_datasets})
         \item Additional Implementation details (Section~\ref{appendix:additional_implementation_details})
    \item Detailed experimental results (Section~\ref{appendix:additional_tables})

\end{itemize}

\section{Training Datasets}
\label{appendix:pretraining_datasets}
As mentioned in Sec. \ref{lab:medical_datasets_collection}, we collect both uni-modal and multi-modal data for the development of the UniMed pretraining dataset. In this section, we provide details about these datasets.
\subsection{Image-Text Datasets}
\noindent \textbf{MIMIC-CXR \cite{johnson2019mimic}:} This dataset is a cornerstone for multimodal learning in the radiology domain. It is a publicly available collection of chest radiographs, consisting of 377,110 images and 227,827 study reports. For pre-training, we utilize the training split of this dataset consisting of \~270 k images. Additionally, following the XrayGPT \cite{thawakar2024xraygpt} framework, we directly utilize the pre-processed summaries of the radiology reports as captions to form the x-ray image-text pairs in MIMIC-CXR.

\noindent \textbf{PMC-OA \cite{lin2023pmc}:} It is a large-scale collection comprising 1.65 million image-text pairs collected from scientific documents in the PubMed Central Open Access (PMC-OA) subset. This dataset was created by extracting and filtering medical figures and captions from over 2.4 biomedical articles, resulting in a comprehensive collection of diverse medical images and their associated captions.

\noindent \textbf{Quilt \cite{ikezogwo2024quilt}:} Quilt is a histopathology-specific dataset consisting of 419,780 images aligned with 768,826 text pairs, curated from publicly available educational YouTube histopathology content. Quilt does not overlap with any existing open-access histopathology datasets, and can be synergistically integrated with various available data sources.

\noindent \textbf{LLaVA-Med \cite{li2024llava}:} We used two datasets from the LLaVA-Med open dataset collection: LLaVA-Med Instruction Tuning and LLaVA-Med Alignment datasets. The instruction tuning dataset, LLaVA-Med 60k contains 60,000 samples selected from five primary imaging modalities—chest X-ray, CT, MRI, histopathology, and gross pathology. We used for pretraining the first version of this dataset which includes inline mentions as additional context and it contains a total of around 265k image-text pairs. We also used the LLaVA-Med 500K dataset, which consists of 500,000 image-text pairs from various medical modalities.

\noindent \textbf{ROCOv2 \cite{ruckert2024rocov2}:} An expanded version of the ROCO (Radiology Objects in Context) dataset, ROCOv2 is a multimodal dataset comprising 79,789 radiological images with captions and medical concepts, sourced from the PMC Open Access Subset. We utilized its training set, containing approximately 61k image-caption pairs, for pretraining purposes.

\noindent \textbf{OpenI \cite{demner2016preparing}:} The OpenI dataset is a collection of chest X-ray images from the Indiana University hospital network, consisting of 6,459 images and 3,955 reports. Following \cite{thawakar2024xraygpt}, we utilized 3,403 high-quality summaries generated from the text reports, along with the corresponding images, for pretraining purposes.

\noindent \textbf{MM-Retinal \cite{wu2024mm}:} MM-Retinal is a multi-modal dataset comprising over 4k high-quality image-text pairs collected from professional fundus diagram books. It includes images from three key sub-modalities: color fundus photography (CFP), fundus fluorescein angiography (FFA), and optical coherence tomography (OCT). We utilized the training set of 2168 images in UniMed.

 By curating and combining datasets from radiology, ophthalmology, pathology, and other medical fields, we ensure broad generalization and robustness across zero-shot as well as downstream adaptation-based medical imaging tasks.
 
 \subsection{Image Only Datasets}
We also collect uni-modal image-only datasets for the devlopement of UniMed. While they only offer images associated with categorical labels, we note that such source of data have enormous potential if processed correctly for multimodal applications. Below we present main image-label datasets used in our work.
 
\noindent \textbf{RadImageNet \cite{mei2022radimagenet}:}
It is a large-scale radiology-specific dataset containing over 1.35 million labeled medical images from diverse imaging modalities such as CT, MRI, ultrasound (US), and more. It covers 157 radiological categories, encompassing various anatomical structures, diseases, and conditions, enabling robust model development across different clinical contexts.

\noindent \textbf{CheXpert \cite{irvin2019chexpert}:} A widely-used chest X-ray dataset containing over 224,000 labeled images from 65,000 patients designed for thoracic disease classification. The dataset provides labels for 14 different pathologies, but the corresponding radiology reports are not available, limiting the dataset to image-only use cases. Despite this, CheXpert remains a key resource for developing models for disease detection and classification in chest radiographs. We used its training set in UniMed.

\noindent \textbf{Flair \cite{silva2024foundation}:} Several retinal fundus image datasets from various open-access sources are collected in  \cite{silva2024foundation} to train a retina modality-specific  foundational model. We utilized more than 25 datasets from this collection to create UniMed dataset. The details of the retinal datasets from Flair used for UniMed development are listed in Table \ref{tab:flairdata}.

\noindent \textbf{Chest X-ray 8 \cite{wang2017chestx}:} The NIH Chest X-ray 8 dataset consists of 112,120 X-ray images with disease labels from 30,805 unique patients. We note that the original radiology reports are not publicly accessible. The authors have provided the labels which were generated using NLP techniques to extract disease classifications from the corresponding radiological reports. 
\begin{table}[ht]
\centering
\small
\renewcommand{\arraystretch}{1.5} 
\begin{tabular}{|l|c|c|}
\hline
\rowcolor{gray!20} \textbf{Dataset} & \textbf{\#Classes} & \textbf{\#Images}  \\
\hline
\rowcolor{gray!10} EYEPACS \cite{cuadros2004eyepacs} & 5 & 3375  \\
IDRID \cite{porwal2020idrid} & 10 & 467  \\
\rowcolor{gray!10} RFMid \cite{pachade2021retinal} & 46 & 2935 \\
LAG \cite{li2019attention} & 2 & 4854 \\
\rowcolor{gray!10} ODIR-5K \cite{bhati2023discriminative} & 2 & 10029 \\
PAPILA \cite{kovalyk2022papila} & 2 & 141 \\
\rowcolor{gray!10} PARAGUAY \cite{castillo2021dataset} & 7 & 757 \\
STARE \cite{hoover2000locating}, \cite{hoover2003locating}  & - & 395 \\
\rowcolor{gray!10} ARIA \cite{farnell2008enhancement} & 3 & 129 \\
AGAR300 \cite{derwin2020novel} & 2 & 28 \\
\rowcolor{gray!10} APTOS & 5 & 3662  \\
FUND-OCT \cite{hassan2019deep}, \cite{hassan2020rag} & 7 & 179 \\
\rowcolor{gray!10} DiaRetDB1 \cite{kauppi2007diaretdb1} & 9 & 86 \\
DRIONS-DB \cite{carmona2008identification} & 1 & 110 \\
\rowcolor{gray!10} Drishti-GS1 \cite{sivaswamy2014drishti} & 2 & 101  \\
E-ophta \cite{decenciere2013teleophta} & 2 & 463 \\
\rowcolor{gray!10} G1020 \cite{bajwa2020g1020} & 2 & 1020  \\
HRF \cite{budai2013robust} & 4 & 45 \\
\rowcolor{gray!10} ORIGA \cite{zhang2010origa} & 2 & 650 \\
ROC \cite{niemeijer2009retinopathy} & 1 & 100  \\
\rowcolor{gray!10} BRSET \cite{nakayama2023brazilian}, \cite{goldberger2000physiobank} & 24 & 15998  \\
OIA-DDR \cite{li2019diagnostic} & 9 & 12522 \\
\rowcolor{gray!10} AIROGS \cite{de2023airogs} & 2 & 101,442 \\
SYSU \cite{lin2020sustech} & 8 & 1219 \\
\rowcolor{gray!10} JICHI \cite{takahashi2017applying} & 5 & 9939 \\
CHAKSU \cite{kumar2023chakṣu} & 2 & 1344 \\
\rowcolor{gray!10} DR1-2 \cite{pires2014advancing} & 7 & 2013 \\
Cataract \cite{silva2024foundation} & 4 & 401 \\
\rowcolor{gray!10} ScarDat \cite{wei2019laser} & 2 & 997 \\
\hline
\end{tabular}
\caption{\textbf{Datasets in Flair collection:} Datasets from Flair retinal Fundus Dataset Collection used in UniMed dataset creation.}
\label{tab:flairdata}
\end{table}

\section{Downstream Datasets} 
\label {appendix:Downstream_datasets}
To evaluate the generalizability of our Medical VLMs including UniMed-CLIP, we conducted zero-shot classification experiments across a comprehensive set of 21 unseen datasets spanning six distinct modalities. These datasets encompass various diagnostic tasks, such as disease detection, organ classification, grading, and tumor identification. As summarized in Table \ref{tab:zeroshot datasets}, the evaluation covers multiple medical imaging modalities, including X-ray, CT, MRI, Ultrasound, Histopathology, and Retinal Imaging. We employed accuracy (ACC) for balanced datasets and area under the curve (AUC) for imbalanced datasets, ensuring a thorough assessment of the model's performance across diverse tasks and class distributions.

Below, we present details of the datasets from different medical imaging modalities used for downstream zero-shot evaluation.

\noindent \textbf{ChexperT 5x200 CXR \cite{irvin2019chexpert}:} Following \cite{huang2021gloria}, we used the ChexperT 5x200 subset of ChexperT chest X-ray dataset which is a multi-class classification dataset containing 200 exclusively positive images for five classes each: Atelectasis, Cardiomegaly, Edema , Pleural Effision, and Pneumonia. 

\noindent \textbf{RSNA Pneumonia CXR \cite{shih2019augmenting}:} This is a binary classification chest X-ray dataset that differentiates between pneumonia and normal cases. Following the approach in \cite{wang2022medclip}, we sampled a balanced subset of 3538 images from the training set, maintaining a 1:1 positive-to-negative ratio for our evaluation purposes. 

\noindent \textbf{MediMeTA Abdomen CT datasets \cite{woerner2024comprehensive}:} The MediMeTA dataset collection includes 19 publicly available datasets covering different modalities. For our evaluation, we used 3 CT datasets from this collection: \textbf{CT-Axial:} Cropped axial slices of 11 abdominal organs from the LiTS dataset \cite{bilic2023liver}. \textbf{CT-Coronal:} Coronal views of the same 11 organs. \textbf{CT-Sagittal:} Sagittal views of the same organs. We followed the same train:test split provided by MediMeTA .

\noindent \textbf{Thyroid Ultrasound \cite{pedraza2015open}:} This dataset is used for binary classification of thyroid nodules in ultrasound images, distinguishing between malignant (288 images) and benign (61 images) cases.

\noindent \textbf{Breast Ultrasound \cite{al2020dataset}:} This binary classification dataset focuses on breast lesion detection using ultrasound, with 210 malignant images and 570 benign images.

\noindent \textbf{Knee MRI \cite{bien2018deep}:} This dataset includes MRI images for two tasks: \textbf{ACL Tear:} A binary classification with 570 ACL tear images and 452 non-ACL tear images, and \textbf{Meniscus Tear:} A binary classification with 506 meniscal tear images and 3695 non-meniscal tear images.

For the Thyroid US, Breast US, ACL, and Meniscus datasets, we applied the same 75:10:15 train:validation:test split as used in \cite{mei2022radimagenet} for both zero-shot and linear probing experiments.

\noindent \textbf{PatchCamelyon (PCam) \cite{veeling2018rotation}:} This binary classification dataset consists of 327,680 color images from histopathology scans of lymph node sections to differentiate the presence or absence of metastatic tissue in breast cancer histopathology slides. We used the official train:test split for zero-shot classification and linear probing experiments. 

\noindent \textbf{LC25000 \cite{borkowski2019lung}:} It consists of (i) \textbf{LC25000 Lung:} This three-class (lung adenocarcinomas, lung squamous cell carcinomas, and benign lung tissue) classification dataset contains 15,000 images  from lung histopathology slides and  (ii) \textbf{LC25000 Colon:} This binary classification (colon adenocarcinomas and benign colonic tissues) dataset contains 10,000 histopathology images of colon tissue. Following \cite{zhang2023biomedclip} and \cite{ikezogwo2024quilt}, we evaluated our trained models on the whole set of images for both datasets.

\noindent \textbf{Skin Cancer \cite{kriegsmann2022deep}:} This dataset consists of 36,890 patches (395×395 pixels) from skin biopsies, used for a 16-class classification task.

\noindent\textbf{Skin Tumor:} A four-class subset of Skin Cancer dataset focuses on differentiating types of skin tumors. Similar to \cite{ikezogwo2024quilt}, we used the official splits to evaluate the trained models.

\noindent \textbf{Osteosarcoma \cite{arunachalam2019viable}:} This dataset contains 1,144 patches from 40 whole-slide images (WSIs) representing the heterogeneity of osteosarcoma. The task is a three-class classification: viable tumor, non-tumor, and necrotic tumor. Following \cite{ikezogwo2024quilt}, we used the entire set of images for testing. As per the approach in \cite{ikezogwo2024quilt}, we utilized the full set of images for testing.

\noindent \textbf{BACH \cite{aresta2019bach}:} This dataset includes 400 histopathology patches of breast tissue, categorized as normal, benign, in-situ carcinoma, and invasive carcinoma (four classes). 

\noindent \textbf{NCT-CRC-HE-100K \cite{kather2018100}:} This dataset comprises 100k non-overlapping image patches from H\&E stained histological images of colorectal cancer. It is categorized into cancerous and normal tissue types for an eight-class classification task. We used the available official test set for evaluation.

\noindent \textbf{MediMeTA Diabetic Retinopathy (Regular Fundus) \cite{woerner2024comprehensive}:}  DR\_Regular dataset  consists of fundus photography images from the DeepDRiD dataset \cite{liu2022deepdrid}, including patients with and without diabetic retinopathy. Diabetic retinopathy grading (ordinal regression) task has five labels. We use the official train test split for experiments.

\noindent \textbf{REFUGE \cite{orlando2020refuge}:} REFUGE is a binary classification dataset for glaucoma detection which consists of total 1200 images. And following \cite{silva2024foundation}, we used the official training-testing split for evaluation and linear probing experiments.

\noindent \textbf{FIVES \cite{jin2022fives}:} It is a fundus image dataset used for the classification of heterogeneous eye diseases. Following \cite{silva2024foundation}, we used the same set of images for evaluation.

\noindent \textbf{ODIR 200x3 \cite{bhati2023discriminative}:} 
It is a subset of the ODIR5 K dataset, which consists of 200 selected images from 3 classes, including normal, CAT, and MYA. 
\begin{table}[t!]
\centering
\small
\renewcommand{\arraystretch}{1.5} 
\begin{tabular}{|l|l|c|l|}
\hline
\rowcolor{gray!20} \textbf{Modality} & \textbf{Dataset} & \textbf{\#Classes} & \textbf{Metric} \\ \hline
\rowcolor{gray!10} \multirow{2}{*}{\textbf{X-ray}} 
    & ChexPerT 5x200 & 5 & ACC \\ 
    & RSNA & 2 & ACC \\ \hline
\multirow{3}{*}{\textbf{CT}} 
    & MediMeTA CT-Axial & 11 & ACC \\ 
    & MediMeTA CT-Coronal & 11 & ACC \\ 
    & MediMeTA CT-Sagittal & 11 & ACC \\ \hline
\rowcolor{gray!10} \multirow{2}{*}{\textbf{MRI}} 
    & ACL & 2 & AUC \\ 
    & Meniscus & 2 & AUC \\ \hline
\multirow{2}{*}{\textbf{US}} 
    & Thyroid & 2 & AUC \\ 
    & Breast & 2 & AUC \\ \hline
\rowcolor{gray!10} \multirow{8}{*}{\textbf{Pathology}} 
    & PCam & 2 & ACC \\ 
    & LC25000 (Lung) & 3 & ACC \\ 
    & LC25000 (Colon) & 2 & ACC \\ 
    & Skin Cancer & 16 & ACC \\ 
    & Skin Tumor & 4 & ACC \\ 
    & Osteosarcoma & 3 & ACC \\ 
    & BACH & 4 & ACC \\ 
    & NCT-CRC & 8 & ACC \\ \hline
\multirow{4}{*}{\textbf{Fundus}} 
    & DR\_Regular & 5 & ACC \\ 
    & Refuge & 2 & AUC \\ 
    & Fives & 6 & ACC \\ 
    & ODIR 200x3 & 3 & ACC \\ \hline
\end{tabular}
\caption{\textbf{Downstream datasets used:} Modality, datasets, the number of classes, and evaluation metrics used for downstream zero-shot evaluation.}
\label{tab:zeroshot datasets}
\end{table}

 \begin{figure*}
  \centering
    \includegraphics[width=0.8\linewidth]{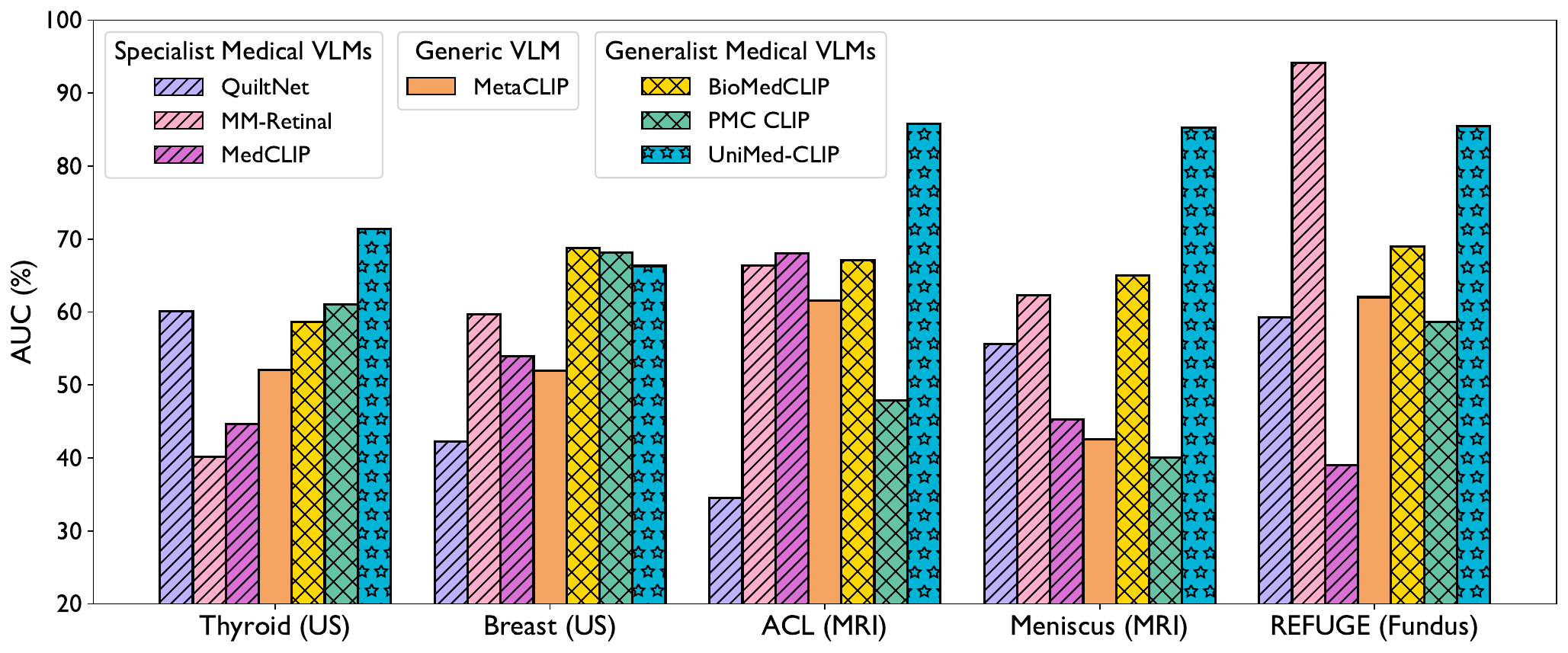}
    \caption{\textbf{Zero-shot performance across class imbalanced datasets:} UniMed-CLIP outperforms other generalist medical VLMs in most cases.}
    \label{fig:AUC_Datasets_plot_plain}
\end{figure*}

\begin{table*}
    \centering
    \resizebox{\textwidth}{!}{ 
    \begin{tabular}{lcccccccccc}
        \toprule
        & \multicolumn{2}{c}{X-ray} & \multicolumn{2}{c}{Ultrasound} & \multicolumn{2}{c}{MRI} & \multicolumn{3}{c}{CT} \\
        \cmidrule(lr){2-3} \cmidrule(lr){4-5} \cmidrule(lr){6-7} \cmidrule(lr){8-10} 
        & CheXpert(5x200) & RSNA & Thyroid & Breast & ACL & Meniscus & Axial & Coronal & Sagittal \\
        \midrule
         \multicolumn{10}{c}{\underline{\textbf{Specialist Models}}} \\
        Quilt 1M (Histopathology) & 18.20 & 45.45 & 60.12 & 42.27 & 34.49 & 55.57 & 10.03 & 5.83 & 9.38 \\
        MM-Retinal (Fundus) & 23.30 & 50.00 & 40.12 & 59.72 & 66.32 & 62.26 & 8.74 & 12.46 & 9.06 \\
        MedCLIP (X-ray) & \underline{59.42} & \underline{73.63} & 44.68 & 53.98 & \underline{68.01} & 45.27 & 9.87 & 13.92 & 10.36 \\
        \midrule
        \multicolumn{10}{c}{\underline{\textbf{Generalist Models}}} \\
        MetaCLIP & 20.10 & 45.25 & 52.05 & 51.92 & 61.54 & 42.60 & 13.43 & 11.97 & 14.24 \\
        PMC CLIP & 30.30 & \textbf{74.99} & 58.60 & \textbf{68.76} & 67.09 & \underline{65.03} & \underline{33.98} & \underline{20.23} & \underline{23.30} \\
        BioMedCLIP & 38.30 & 72.56 & \underline{61.05} & \underline{68.12} & 47.89 & 40.09 & 29.13 & 19.09 & 20.39 \\
        \midrule
      \rowcolor{cyan!15}  UniMed-CLIP & \textbf{65.90} & 71.65 & \textbf{71.35} & 66.31 & \textbf{85.82} & \textbf{85.27} & \textbf{37.54} & \textbf{24.43} & \textbf{31.72} \\
        \bottomrule
    \end{tabular}
    }
    \caption{\textbf{Zero-shot classification performance on various radiology datasets.} Best results are highlighted in bold and second-best results are underlined.}
    \label{tab:radiology-results} 
\end{table*}

\begin{table*}
    \centering
    \resizebox{\textwidth}{!}{  
    \begin{tabular}{l c c c c c c c c||c}
        \hline
        \textbf{Model} & \textbf{LC25000\_lung} & \textbf{LC25000\_colon} & \textbf{PCAM} & \textbf{BACH} & \textbf{NCT-CRC} & \textbf{SkinCancer} & \textbf{SkinTumor} & \textbf{Osteo} & \textbf{Avg ACC} \\
        \hline\hline
        \multicolumn{10}{c}{\underline{\textbf{Specialist Models}}} \\
        \hline
        Quilt 1M & \textbf{80.17} & \underline{93.93} & \textbf{64.43} & 41.60 & \textbf{59.58} & \textbf{45.38} & \textbf{58.29} & \underline{63.52} & \textbf{63.11} \\
        MM-Retinal & 45.75 & 25.82 & 53.23 & 24.06 & 18.44 & 6.41 & 1.98 & 39.05 & 26.09 \\
        MedCLIP & 31.91 & 50.83 & 50.24 & 23.06 & 8.18 & 8.49 & 30.05 & 26.12 & 28.61 \\  
        \hline
        \multicolumn{10}{c}{\underline{\textbf{Generalist Models}}} \\
        \hline
        MetaCLIP & 66.41 & 65.03 & 51.79 & 33.83 & 32.91 & 15.18 & 33.11 & 20.35 & 39.08 \\
        PMC CLIP & 70.80 & 69.46 & \underline{58.69} & 43.85 & \underline{58.60} & 30.91 & 47.81 & 48.40 & 53.19 \\
        BioMedCLIP & 72.00 & 91.00 & 53.00 & \textbf{55.00} & 52.00 & 25.00 & 37.00 & 51.00 & 54.25 \\
        \hline
        \rowcolor{cyan!15} UniMed-CLIP & \underline{79.99} & \textbf{94.87} & 52.42 & \underline{54.38} & 49.83 & \underline{32.05} & \underline{49.87} & \textbf{66.26} & \underline{59.96} \\
        \hline
    \end{tabular}
    }
    \caption{\textbf{Zeroshot evaluation across eight histopathology datasets}. Best results are highlighted in bold and second-best results are underlined. On average, UniMed CLIP outperforms other generalist Medical VLMs, and fares competitive to Quilt-1M Histopathology specialist VLM.}
    \label{tab:histopathology-results} 
\end{table*}

 \section{Additional Implementation details}
 \label{appendix:additional_implementation_details}

\textbf{UniMed-CLIP Training:} For UniMed-CLIP pretraining, we initialize its vision-encoder from MetaCLIP ViT-B/16 model \cite{xu2023demystifying}, and its text encoder is initialized from the BioMed-BERT text encoder \cite{chakraborty2020biomedbert} respectively. We fine-tune UniMed-CLIP using per GPU batch size of 128 (with an effective batch size of 2048) for 10 epochs using 16 A100 40GB GPUs in a multi-node training setup. The learning rate is set to 5e-5 with a warmup phase of 2k iterations. Complete training of UniMed-CLIP takes 10 hours in total. For pretraining UniMed-CLIP, we adopt the training code-base of Meta-CLIP \cite{xu2023demystifying}.

\begin{figure*}
  \centering
     \includegraphics[width=\linewidth]{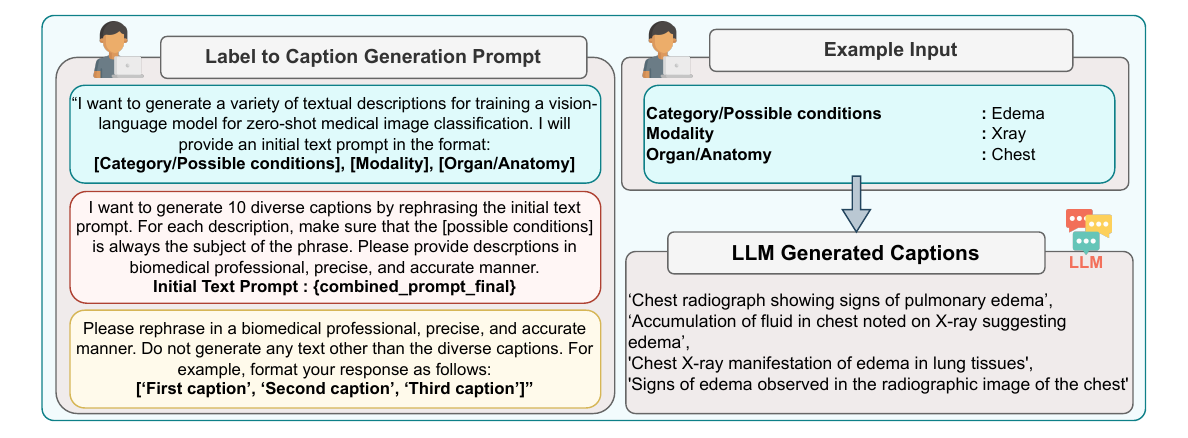}
  \caption{\textbf{LLM prompt message for generating captions for uni-modal datasets.} We provide detailed instruction to LLM alongside anatomy, modality, and label/disease information to generate template captions consistent with professional medical style and tone.}
  \label{appendix: fig:sample_prompt}
\end{figure*}

\begin{figure*}
  \centering
     \includegraphics[width=\linewidth]{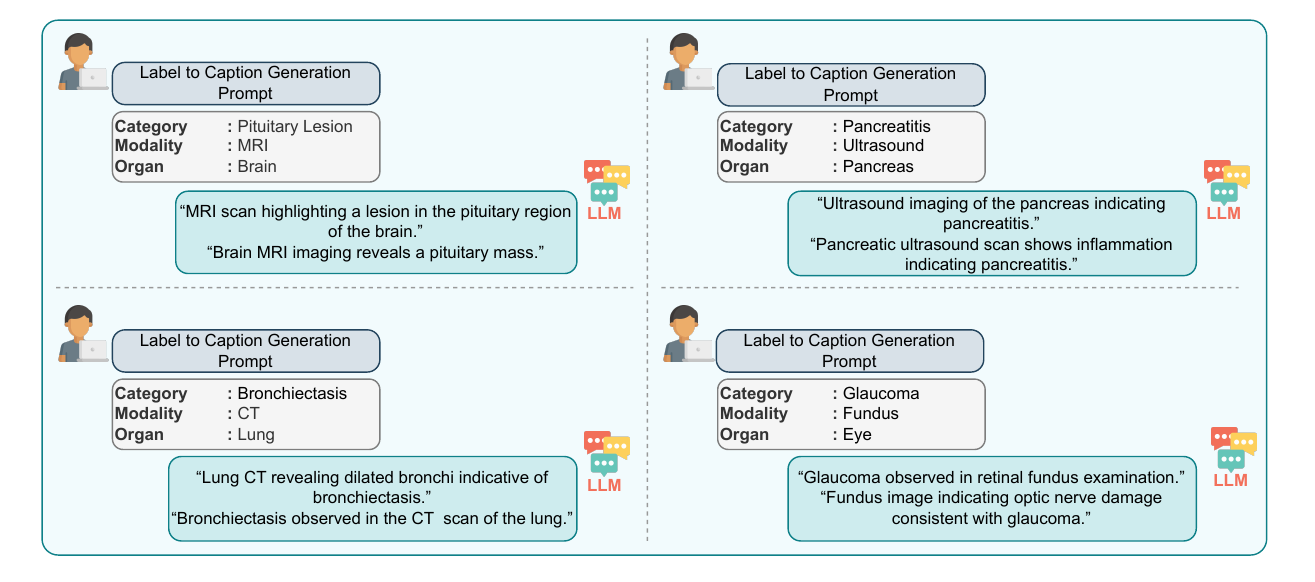}
  \caption{\textbf{Qualitative examples for LLM generated captions:} We show qualitative examples for template captions generated for various medical modalities.}
  \label{appendix: fig:example_captions}
\end{figure*}

\noindent\textbf{Label to Template Caption Generation:}
In order to convert label-only (uni-modal) datasets into vision-language multi-modal format, we formulate the label-to-template caption as described in Sec. \ref{lab:medical_datasets_collection_label_only} the main paper. For the choice of LLM, we utilize GPT-4o \cite{gpt4o} to generate captions. An example prompt message to LLM is shown in Fig. \ref{appendix: fig:sample_prompt}. Additional qualitative results for template caption are shown in Fig. \ref{appendix: fig:example_captions}.

 \section{Detail Experimental Results}
 \label{appendix:additional_tables}
\textbf{Zero-shot Experiments.} Here, we provide per-dataset evaluation results for zero-shot.  For zero-shot radiology datasets evaluation, results are shown in Tab. \ref{tab:radiology-results}. For Histopathology, we present the comparisons of the results in Tab. \ref{tab:histopathology-results}. 

\noindent \textbf{Linear Probing:} We present the per-dataset linear probing results in Tab. \ref{appendix:linear_probing_table}. We compare UniMed-CLIP with prior generalist medical VLMs on 1\%, 10\% and 100\% data points.
 
\noindent \textbf{Results on Class-Imbalanced Datasets:} Also, as shown in Fig. \ref{fig:AUC_Datasets_plot_plain}, our model maintains strong performance across class-imbalanced datasets from various modalities, including CT, MRI, and Retinal fundus imaging. These results underscore the versatility of the UniMed-CLIP, highlighting its ability to generalize well even in tasks where class distributions are skewed.

\begin{table*}[ht]
\centering
\begin{tabular}{c|c|c c c}
\hline
\textbf{Dataset} &  & \multicolumn{3}{c}{\textbf{Performance \%}} \\
\cline{3-5} 
(Modality) & \textbf{Model} & \textbf{1\%} & \textbf{10\%} & \textbf{100\%} \\ \hline \hline
\multirow{4}{*}{PCam } & CLIP & 79.97 & 82.99 & 83.87 \\ 
                   & PMC-CLIP & 78.45 & 79.38 & 81.03 \\ 
                   & BiomedCLIP & 83.23 & 83.64 & 83.40 \\ 
   (Histopathology) & \cellcolor{cyan!15} UniMed-CLIP & \cellcolor{cyan!15} \textbf{84.66} & \cellcolor{cyan!15} \textbf{85.59} & \cellcolor{cyan!15} \textbf{85.97} \\ \hline
\multirow{4}{*}{Skin Cancer} & CLIP & 72.47 & 80.80 & 86.70 \\ 
                              & PMC-CLIP & 6.76 & 5.57 & 11.09 \\ 
                              & BiomedCLIP & 71.21 & 84.58 & 87.39 \\ 
    (Histopathology) & \cellcolor{cyan!15} UniMed-CLIP & \cellcolor{cyan!15} \textbf{82.10} & \cellcolor{cyan!15} \textbf{86.83} & \cellcolor{cyan!15} \textbf{88.97} \\ \hline
\multirow{4}{*}{RSNA} & CLIP & 71.67 & 73.89 & 81.09 \\ 
                      & PMC-CLIP & 65.25 & 64.15 & 79.47 \\ 
                      & BiomedCLIP & \textbf{80.04} & 78.32 & 83.84 \\ 
    (X-ray) & \cellcolor{cyan!15} UniMed-CLIP & \cellcolor{cyan!15} 79.52 & \cellcolor{cyan!15} \textbf{79.19} & \cellcolor{cyan!15} \textbf{88.51} \\ \hline
\multirow{4}{*}{Thyroid } & CLIP & 70.99 & 72.51 & 75.99 \\ 
                          & PMC-CLIP & 46.78 & 56.61 & 61.40 \\ 
                          & BiomedCLIP & \textbf{76.96} & \textbf{80.47} & \textbf{81.87} \\ 
    (US) & \cellcolor{cyan!15} UniMed-CLIP & \cellcolor{cyan!15} 55.67 & \cellcolor{cyan!15} 64.44 & \cellcolor{cyan!15} 76.84 \\ \hline
\multirow{4}{*}{Breast } & CLIP & \textbf{64.38} & 76.52 & 77.94 \\ 
                         & PMC-CLIP & 50.59 & 61.08 & 57.50 \\ 
                         & BiomedCLIP & 63.92 & \textbf{75.59} & \textbf{82.90} \\ 
    (US) & \cellcolor{cyan!15} UniMed-CLIP & \cellcolor{cyan!15} 49.29 & \cellcolor{cyan!15} 67.09 & \cellcolor{cyan!15} 77.22 \\ \hline
\multirow{4}{*}{ACL } & CLIP & 56.99 & 85.89 & 91.73 \\ 
                      & PMC-CLIP & 57.00 & 63.47 & 77.92 \\ 
                      & BiomedCLIP & 44.13 & 65.46 & 90.12 \\ 
    (MRI) & \cellcolor{cyan!15} UniMed-CLIP & \cellcolor{cyan!15} \textbf{89.61} & \cellcolor{cyan!15} \textbf{95.28} & \cellcolor{cyan!15} \textbf{97.28} \\ \hline
\multirow{4}{*}{Meniscus } & CLIP & 74.87 & 79.99 & 83.59 \\ 
                           & PMC-CLIP & 58.77 & 60.49 & 62.31 \\ 
                           & BiomedCLIP & 77.46 & 82.52 & 84.99 \\ 
    (MRI) & \cellcolor{cyan!15} UniMed-CLIP & \cellcolor{cyan!15} \textbf{90.52} & \cellcolor{cyan!15} \textbf{92.94} & \cellcolor{cyan!15} \textbf{94.20} \\ \hline
\multirow{4}{*}{MediMeTA Axial} & CLIP & 32.20 & 45.97 & 70.06 \\ 
                                & PMC-CLIP & 35.92 & 43.04 & 57.61 \\ 
                                & BiomedCLIP & 29.77 & 59.06 & \textbf{77.67} \\ 
    (CT) & \cellcolor{cyan!15} UniMed-CLIP & \cellcolor{cyan!15} \textbf{39.97} & \cellcolor{cyan!15} \textbf{57.28} & \cellcolor{cyan!15} 76.38 \\ \hline
\multirow{4}{*}{MediMeTA Coronal} & CLIP & 17.96 & 28.32 & 52.59 \\ 
                                  & PMC-CLIP & 21.20 & 34.63 & 55.02 \\ 
                                  & BiomedCLIP & \textbf{30.26} & 45.95 & 61.00 \\ 
    (CT) & \cellcolor{cyan!15} UniMed-CLIP & \cellcolor{cyan!15} 29.94 & \cellcolor{cyan!15} \textbf{48.06} & \cellcolor{cyan!15} \textbf{65.70} \\ \hline
\multirow{4}{*}{MediMeTA Sagittal} & CLIP & 21.20 & 34.14 & 48.06 \\ 
                                   & PMC-CLIP & 27.83 & 38.19 & 47.73 \\ 
                                   & BiomedCLIP & 29.45 & 44.66 & 56.31 \\ 
    (CT) & \cellcolor{cyan!15} UniMed-CLIP & \cellcolor{cyan!15} \textbf{36.57} & \cellcolor{cyan!15} \textbf{54.37} & \cellcolor{cyan!15} \textbf{62.78} \\ \hline
\multirow{4}{*}{REFUGE } & CLIP & 64.76 & 74.31 & 76.04 \\ 
                         & PMC-CLIP & 41.49 & 51.22 & 63.02 \\ 
                         & BiomedCLIP & 62.67 & 65.45 & 75.00 \\ 
    (Fundus) & \cellcolor{cyan!15} UniMed-CLIP & \cellcolor{cyan!15} \textbf{82.29} & \cellcolor{cyan!15} \textbf{88.19} & \cellcolor{cyan!15} \textbf{89.24} \\ \hline
\multirow{4}{*}{ODIR 2x300 } & CLIP & 88.33 & 89.17 & 91.67 \\ 
                              & PMC-CLIP & 33.33 & 65.00 & 70.83 \\ 
                              & BiomedCLIP & \textbf{89.17} & 94.17 & 94.17 \\ 
    (Fundus) & \cellcolor{cyan!15} UniMed-CLIP & \cellcolor{cyan!15} 86.67 & \cellcolor{cyan!15} \textbf{95.00} & \cellcolor{cyan!15} \textbf{95.00}\\ \hline
\multirow{4}{*}{FIVES } & CLIP & 31.87 & 69.38 & 76.25 \\ 
                        & PMC-CLIP & 20.62 & 55.62 & 57.50 \\ 
                        & BiomedCLIP & 30.00 & 58.75 & 66.88 \\ 
    (Fundus) & \cellcolor{cyan!15} UniMed-CLIP & \cellcolor{cyan!15} \textbf{43.12} & \cellcolor{cyan!15} \textbf{76.88} & \cellcolor{cyan!15} \textbf{78.75} \\ \hline
\end{tabular}
\caption{Linear Probing Results: Performance comparison across different generalist medical foundation models with varying training data percentages.}
\label{appendix:linear_probing_table}
\end{table*}

\end{document}